\documentclass[11pt]{article}

\usepackage[margin=0.9in]{geometry}
\usepackage[utf8]{inputenc}
\usepackage[T1]{fontenc}
\usepackage{lmodern}
\usepackage{microtype}
\usepackage{amsmath,amssymb,amsfonts,bm}
\usepackage{booktabs}
\usepackage{multirow}
\usepackage{array}
\usepackage{tabularx}
\usepackage{makecell}
\usepackage{graphicx}
\usepackage{xcolor}
\usepackage{enumitem}
\usepackage{natbib}
\usepackage{caption}
\usepackage{subcaption}
\usepackage{algorithm}
\usepackage{algpseudocode}
\usepackage{float}
\usepackage{rotating}
\usepackage{tikz}
\usetikzlibrary{arrows.meta,positioning,fit}
\usepackage{hyperref}

\hypersetup{
    colorlinks=true,
    linkcolor=blue!55!black,
    citecolor=blue!55!black,
    urlcolor=blue!55!black,
    pdftitle={qZACH-ViT: Quantization-Aware Intrinsic Explanations with Recursive Attribution-Stabilized Optimization},
    pdfauthor={Athanasios Angelakis},
    pdfsubject={Quantization-aware compact Vision Transformers with intrinsic explanations, attribution-stabilized optimization, and validated ONNX INT8 deployment},
    pdfkeywords={qZACH-ViT, ZACH-ViT, RASO, quantization-aware training, ONNX INT8, intrinsic explainability, Vision Transformer, MedMNIST}
}

\graphicspath{{figures/}}
\setlist[itemize]{leftmargin=*,topsep=3pt,itemsep=2pt}
\setlist[enumerate]{leftmargin=*,topsep=3pt,itemsep=2pt}
\newcommand{\qzach}{qZACH-ViT}
\newcommand{\zach}{ZACH-ViT}
\newcommand{\raso}{RASO}
\newcommand{\Lcls}{\mathcal{L}_{\mathrm{cls}}}
\newcommand{\Lattr}{\mathcal{L}_{\mathrm{attr}}}
\newcommand{\JS}{\operatorname{JS}}

\title{qZACH-ViT: Quantization-Aware Intrinsic Explanations with Recursive Attribution-Stabilized Optimization}

\author{
Athanasios Angelakis\textsuperscript{1,2}\\[2pt]
\textsuperscript{1}BioML Lab, Research Institute CODE, UniBw, Munich, Germany\\
\textsuperscript{2}Epidemiology and Data Science, Amsterdam UMC, Amsterdam, Netherlands\\
\texttt{athanasios.angelakis@unibw.de}
}

\date{}

\begin{document}
\maketitle

\begin{abstract}
Compact medical-image classifiers need efficiency and interpretable evidence, yet these goals are often addressed separately. We introduce qZACH-ViT, a quantization-aware extension of the zero-token
(CLS-token-free), position-free ZACH-ViT backbone with recursive intrinsic
patch-level class evidence. We also introduce Recursive Attribution-Stabilized
Optimization (RASO), which norm-matches classification and attribution gradients
and removes attribution components that conflict with classification. We evaluate four controlled conditions on seven MedMNIST datasets using 50 training images per class and ten fixed seeds, completing 280 runs. All 210 qZACH-ViT checkpoints are converted to executable mixed-precision ONNX INT8 graphs containing 16 signed INT8 \texttt{MatMulInteger} projections with INT32 accumulation. Deployed mixed-precision INT8 qZACH-ViT with Adam improves the FP32 ZACH-ViT baseline mean on all seven datasets, with a mean paired gain of 0.0313 in the dataset-specific primary metric; qZACH-ViT with RASO yields a mean gain of 0.0368. Across 964,920 source-to-INT8 test comparisons, prediction agreement is 99.9751\%, with a mean absolute primary-metric change of 0.000133 and a maximum of 0.004386. Across 3,600 matched intrinsic maps, mean cosine similarity is 0.999955, mean rank correlation is 0.9944, and mean top-10\% overlap is 0.9692. ONNX artifacts are 70.0\% smaller than source checkpoints and provide $1.41\times$ and $2.39\times$ end-to-end CPU speedups with one and four threads. RASO significantly reduces sufficiency error and improves input-noise stability over Adam with the same attribution loss, but does not dominate every predictive or explainable artificial intelligence (XAI) metric. These results establish qZACH-ViT as a deployable compact intrinsically explainable model and RASO as a targeted stability-oriented optimization procedure.
\end{abstract}

\section{Introduction}
Vision Transformers can be accurate but expensive, and their predictions are commonly explained only after training. These issues become tightly coupled under compression: a quantized model may preserve a class label while changing the internal evidence that supports it. In medical imaging, prediction retention alone is therefore insufficient. Deployment validation should ask whether the converted model remains accurate, whether its evidence remains stable, and whether the claimed integer operations are actually present and executed.

The starting point of this study is the ZACH-ViT line. The architecture was first introduced for low-data lung-ultrasound classification \citep{angelakis2025zachlus}, then formalized and evaluated systematically across seven MedMNIST tasks \citep{angelakis2026zachvit}, and subsequently stress-tested under common corruptions and adversarial perturbations \citep{angelakis2026zachrobust}. The present work addresses a different question: whether the same compact zero-token backbone can provide intrinsic patch evidence, support explanation-stabilized optimization, and survive conversion to an executable INT8 deployment graph. ZACH-ViT uses neither positional embeddings nor a classification token, aggregates patch representations by global average pooling, and contains adaptive residual projections. This design yields approximately 0.25 million parameters and makes direct patch-wise evidence accounting possible without attaching a separate post-hoc explainer.

We develop this opportunity in two steps. The first contribution, \qzach{}, is an architectural and deployment extension. It adds W8A8 quantization-aware training and two patch-level class-evidence heads. The raw evidence exactly reconstructs the class logit through spatial averaging and recursive stage fusion, while normalized maps are available during training and inference. After training, the selected checkpoints are converted to mixed-precision ONNX graphs in which all 16 learned projections use signed INT8 \texttt{MatMulInteger} operators with INT32 accumulation. The second contribution, \raso{}, is an optimization procedure for the coupled classification and attribution objectives. It norm-matches the attribution gradient to the classification gradient and, when the two conflict, removes the conflicting attribution component before the Adam update. Classification is treated as the protected objective, while explanation consistency is optimized when compatible.

The distinction between these contributions determines the controlled experiment. \zach{} with Adam is the full-precision baseline. \qzach{} with Adam isolates quantization-aware intrinsic explanation. \qzach{} with Adam plus the same attribution loss tests whether ordinary scalar loss addition is sufficient. \qzach{} with \raso{} tests the additional value of norm matching and conflict-aware projection. The three \qzach{} conditions share identical initial states within every dataset-seed pair. Predictive results for the three \qzach{} conditions are reported from the actual converted ONNX INT8 models, not from the fake-quantized training path.

Our contributions are:
\begin{itemize}
    \item We introduce \qzach{}, a compact quantization-aware extension of \zach{} with recursive intrinsic patch-level class evidence and exact raw-evidence logit completeness.
    \item We introduce \raso{}, an asymmetric gradient-combination procedure that targets recursive and quantization-consistent explanations while protecting the classification direction.
    \item We validate actual deployment for 210 converted checkpoints. Every graph contains 16 signed INT8 \texttt{MatMulInteger} projections, and every ONNX Runtime profile records integer execution.
    \item We evaluate 964,920 source-to-INT8 test predictions and 3,600 matched intrinsic maps, together with 280 controlled training runs, 21,600 source-model XAI evaluations, and 168 parameter-randomization checks.
    \item We report positive and negative findings. Actual INT8 \qzach{} improves the baseline mean on all seven datasets and retains predictions, primary metrics, and intrinsic maps with high fidelity. \raso{} strengthens selected stability and sufficiency properties, but it is not universally best in prediction or faithfulness, and several post-hoc methods remain stronger on individual XAI metrics.
\end{itemize}

The deployment is mixed precision rather than fully integer-only. LayerNorm, residual additions, attention score and context products, softmax, pooling, dequantization, bias addition, and attribution normalization remain in FP32. The study also contains no localization annotations and makes no clinical-localization claim.

\section{Related Work}
\paragraph{Compact Vision Transformers and the ZACH-ViT lineage.}
The original Vision Transformer represents an image as a sequence of patch tokens and generally benefits from large-scale pretraining \citep{dosovitskiy2020vit}; DeiT showed that careful distillation and augmentation can improve data efficiency \citep{touvron2021deit}. ZACH-ViT follows a different compact-design route. It was first introduced for lung-ultrasound classification \citep{angelakis2025zachlus}, then formalized as a position-free, zero-token architecture and evaluated under a strict low-data MedMNIST protocol \citep{angelakis2026zachvit}. A later study examined its corruption and adversarial robustness \citep{angelakis2026zachrobust}. The current work is not a robustness extension. It introduces a quantization-aware intrinsically explainable branch, qZACH-ViT, and a dedicated optimization method, RASO.

\paragraph{Quantization of Vision Transformers.}
Quantization-aware training uses simulated low-precision operations during optimization so that a model can adapt to clipping and rounding \citep{jacob2018quantization}; broader treatments distinguish fake-quantized training from converted integer execution and hardware realization \citep{gholami2022survey}. Vision Transformers create additional challenges because LayerNorm, GELU, softmax, attention maps, and activation outliers are not uniformly quantization-friendly. Prior work includes fully differentiable learned-bit quantization \citep{li2022qvitdiff}, post-training twin-uniform quantization \citep{yuan2022ptq4vit}, fully quantized LayerNorm and softmax approximations \citep{lin2022fqvit}, integer-only ViT inference \citep{li2023ivit}, low-bit attention rectification and distillation \citep{li2022qvitaccurate}, and scale reparameterization for post-training quantization \citep{li2023repqvit}. Our goal differs from maximizing ImageNet accuracy at the lowest possible bit width. We use a fixed signed W8A8 scheme to study whether decisions and intrinsic patch evidence can be retained together in a compact model. Quantization-aware training is followed by explicit ONNX conversion and execution auditing: all learned projections use signed INT8 matrix multiplication with INT32 accumulation, while non-projection operations remain in FP32.

\paragraph{Post-hoc, intrinsic, and explanation-by-design models.}
Integrated Gradients is derived from sensitivity and implementation-invariance axioms \citep{sundararajan2017ig}. Grad-CAM uses class gradients to weight feature maps \citep{selvaraju2017gradcam}. Attention rollout estimates information flow across transformer layers \citep{abnar2020attention}, while transformer-specific relevance propagation extends explanation beyond raw attention \citep{chefer2021transformer}. RISE estimates salience through randomized masking and black-box prediction changes \citep{petsiuk2018rise}. These methods are useful, but attention alone need not be explanatory \citep{jain2019attention}, explanations may violate input invariance \citep{kindermans2019reliability}, and visually plausible maps can persist after model randomization \citep{adebayo2018sanity}.

Intrinsic approaches make interpretability part of the predictive computation. Self-explaining neural networks learn locally interpretable basis concepts \citep{alvarez2018senn}; prototype networks explain decisions through similarity to learned exemplars \citep{chen2019protopnet}; concept bottleneck models predict human-specified concepts before the target \citep{koh2020cbm}; and B-cos models promote input-weight alignment for inherently interpretable vision models \citep{bohle2023alignment,bohle2023holistic}. qZACH-ViT follows another route: the class logit is explicitly decomposed into patch-level evidence at two recursive depths. The raw evidence is complete by construction, whereas the normalized map is used for visualization and regularization.

\paragraph{Faithfulness, robustness, and evaluation of explanations.}
Deletion and insertion tests measure score changes when regions are removed or restored in attribution order \citep{petsiuk2018rise}. Infidelity and sensitivity formalize complementary desirable properties \citep{yeh2019infidelity}, while remove-and-retrain evaluation tests whether identified features carry predictive information under retraining \citep{hooker2019roar}. Quantus organizes families of faithfulness, robustness, localization, complexity, and randomization metrics \citep{hedstrom2023quantus}. Explanations can nevertheless be fragile under visually negligible perturbations \citep{ghorbani2019interpretation,dombrowski2019explanations}, and deletion-style metrics can be affected by out-of-distribution perturbations and rank-only evaluation \citep{gomez2022metrics}. SaCo was proposed to better discriminate faithful transformer explanations from random attribution \citep{wu2024faithfulness}. We therefore combine perturbation, sufficiency, stability, runtime, and randomization measures rather than treating any single score as decisive. The seven MedMNIST archives contain no masks or boxes, so we do not report localization scores without ground truth.

\paragraph{Gradient conflict and multi-objective optimization.}
Multi-gradient descent seeks Pareto-stationary solutions for multi-task learning \citep{sener2018mgda}. GradNorm balances loss gradients by their magnitudes \citep{chen2018gradnorm}. PCGrad removes conflicting components between task gradients \citep{yu2020pcgrad}, GradVac adapts gradient similarity targets \citep{wang2020gradvac}, CAGrad seeks a conflict-averse direction with multi-objective guarantees \citep{liu2021cagrad}, and Nash-MTL formulates gradient combination as a bargaining problem \citep{navon2022nash}. RASO is related to this literature and is not presented as the first gradient-projection method. Its contribution is the specific asymmetric construction for a protected classification objective and a recursive quantization-consistency objective: it combines norm matching with projection of only the attribution gradient, then applies Adam. The loss-only control determines whether this procedure adds value beyond ordinary scalar regularization.

\section{Method}
\subsection{Backbone}
An RGB image $x\in\mathbb{R}^{224\times224\times3}$ is divided into non-overlapping $16\times16$ patches, producing $N=196$ flattened patch vectors. A learned projection maps each patch to 128 dimensions, followed by ReLU. Two transformer blocks use 8-head self-attention, pre-normalization, dropout 0.1, ReLU feed-forward layers, and adaptive residual projection when the feature dimension changes. The block widths are 128 and 64. The original \zach{} baseline then applies global average pooling, a 128-to-64 MLP head, and the classifier. No positional embeddings and no classification token are used \citep{angelakis2026zachvit}.

\subsection{qZACH-ViT}
\qzach{} preserves the zero-token backbone but changes the prediction head so that class evidence is available at each patch. Let $h_i^{(1)}$ be the representation of patch $i$ after the first transformer block. A linear evidence head produces
\begin{equation}
    e_i^{(1)} = W_1 h_i^{(1)} + b_1.
\end{equation}
After the second transformer block, a token-wise MLP with widths 128 and 64 produces $r_i^{(2)}$, followed by a second evidence head
\begin{equation}
    e_i^{(2)} = W_2 r_i^{(2)} + b_2.
\end{equation}
For multiclass tasks, $e_i^{(r)}\in\mathbb{R}^{C}$. Binary tasks use one logit. Stage logits and the final logit are
\begin{equation}
    z^{(r)} = \frac{1}{N}\sum_{i=1}^{N} e_i^{(r)},
    \qquad
    z = \rho z^{(1)} + (1-\rho)z^{(2)},
    \label{eq:logit_decomposition}
\end{equation}
with $\rho=0.25$. Equation~\ref{eq:logit_decomposition} is an exact decomposition: the implementation audit obtains zero reconstruction error at floating-point precision for all seven output dimensions tested. \qzach{} contains 248,386 to 250,326 parameters depending on class count, only 129 to 1,419 more than the corresponding baseline.

\subsection{Recursive intrinsic attribution}
For target class $c$, raw patch evidence is converted into a normalized map by
\begin{equation}
    A_i^{(r)}(c) =
    \frac{\exp(e_{i,c}^{(r)}/\tau)}
    {\sum_{j=1}^{N}\exp(e_{j,c}^{(r)}/\tau)},
    \qquad \tau=1.
\end{equation}
For binary classification, evidence is signed according to the target label before normalization. The recursive attribution is
\begin{equation}
    A^{\mathrm{rec}}(c) = \rho A^{(1)}(c) + (1-\rho)A^{(2)}(c).
    \label{eq:recursive_map}
\end{equation}
The exact completeness statement applies to the raw evidence in Equation~\ref{eq:logit_decomposition}. The normalized maps in Equation~\ref{eq:recursive_map} preserve patch order and support comparison and visualization, but they are not themselves additive logit decompositions.

\subsection{W8A8 quantization-aware training}
Weights and activations use signed symmetric fake quantization during training with integer range $[-127,127]$ and a straight-through estimator. For tensor $u$ and scale $s$,
\begin{equation}
    Q(u;s) = s\,\operatorname{clip}
    \left(\operatorname{round}(u/s),-127,127\right).
    \label{eq:fake_quant}
\end{equation}
Weight scales use the current absolute maximum per output channel. Activation scales use a per-tensor exponential moving average of the absolute maximum with momentum 0.95. Activation observers are updated during the first three epochs and then frozen. Quantization-aware operations cover the patch projection, query, key, value and attention-output projections, feed-forward layers, adaptive residual projections, token MLP, and class-evidence heads. Biases, LayerNorm, attention softmax, residual additions, recursive fusion, loss computation, Adam states, and matrix accumulation remain in FP32 during training.

Each quantized training forward pass is paired with a float-shadow pass using the same weights, target class, dropout state, and random-number state. This matched path supports the attribution objective below; it is not used as the final deployment result.

\subsection{Exact mixed-precision ONNX INT8 deployment}
After checkpoint selection, the frozen QAT scales and trained weights are converted directly into an ONNX graph. For every learned projection, activations are clipped to the trained signed range and quantized per tensor to INT8. Weights are quantized per output channel to signed INT8. Each projection is evaluated as
\begin{equation}
    y = \operatorname{FP32}\!\left(
    \operatorname{MatMulInteger}(x_{8},W_{8}^{\top})
    \right)\odot(s_x s_W)+b,
    \label{eq:int8_deployment}
\end{equation}
where the matrix product accumulates in INT32, $s_x$ is the frozen activation scale, $s_W$ is the vector of per-output-channel weight scales, and $b$ is added in FP32. Output quantization is applied only where it was present during QAT. The two evidence-head outputs remain FP32 after integer accumulation, matching the trained model rather than introducing an untrained output bottleneck.

Every converted model contains 16 \texttt{MatMulInteger} projection nodes and 16 signed INT8 weight initializers. LayerNorm, residual additions, ReLU, attention score and context products, softmax, pooling, dequantization, bias addition, and attribution normalization remain FP32. The deployment is therefore an actual mixed-precision ONNX INT8 implementation, not a fully integer-only network.

\subsection{Attribution objective}
Let $A_q^{(1)}$, $A_q^{(2)}$, and $A_q^{\mathrm{rec}}$ denote quantized-training-path maps, and let $A_{fp}^{\mathrm{rec}}$ denote the matched float-shadow map. The attribution objective is
\begin{equation}
    \Lattr =
    \frac{1}{2}\JS\left(A_q^{(1)},A_q^{(2)}\right)
    +
    \frac{1}{2}\JS\left(A_q^{\mathrm{rec}},A_{fp}^{\mathrm{rec}}\right).
    \label{eq:attr_loss}
\end{equation}
The first term encourages agreement between recursive depths. The second encourages retention under the training-time quantization model. The loss-only control minimizes $\Lcls+0.1\Lattr$ with ordinary Adam.

\subsection{Recursive Attribution-Stabilized Optimization}
\raso{} computes separate gradients
\begin{equation}
    g_c = \nabla_\theta \Lcls,
    \qquad
    g_a = \nabla_\theta \Lattr.
\end{equation}
The attribution gradient is first norm-matched:
\begin{equation}
    u = \frac{\lVert g_c\rVert_2}
    {\lVert g_a\rVert_2+\epsilon} g_a.
\end{equation}
When $\langle g_c,u\rangle<0$, its conflicting component is removed:
\begin{equation}
    \tilde{u} = u -
    \frac{\langle u,g_c\rangle}
    {\lVert g_c\rVert_2^2+\epsilon}g_c.
    \label{eq:projection}
\end{equation}
Otherwise, $\tilde{u}=u$. The gradient supplied to Adam is
\begin{equation}
    g_{\raso}=g_c+\lambda_a\tilde{u},
    \qquad \lambda_a=0.1.
    \label{eq:raso_gradient}
\end{equation}
This asymmetric choice protects the classification direction: the attribution objective can contribute an orthogonal or aligned component, but it cannot introduce a first-order component directly opposed to $g_c$.

\begin{algorithm}[t]
\caption{One \raso{} optimization step}
\label{alg:raso}
\begin{algorithmic}[1]
\Require Batch $(x,y)$, model parameters $\theta$, coefficient $\lambda_a$
\State Compute quantized and matched float-shadow forward passes
\State Compute $\Lcls$ and $\Lattr$ from Equation~\ref{eq:attr_loss}
\State $g_c\gets\nabla_\theta\Lcls$, $g_a\gets\nabla_\theta\Lattr$
\State $u\gets \lVert g_c\rVert_2 g_a/(\lVert g_a\rVert_2+\epsilon)$
\If{$\langle g_c,u\rangle<0$}
    \State Compute $\tilde{u}$ using Equation~\ref{eq:projection}
\Else
    \State $\tilde{u}\gets u$
\EndIf
\State Supply $g_c+\lambda_a\tilde{u}$ to the Adam moment update
\end{algorithmic}
\end{algorithm}

\section{Experimental Design}
\subsection{Datasets and low-data protocol}
We use seven 2D MedMNIST datasets \citep{yang2023medmnist}: BloodMNIST, PathMNIST, BreastMNIST, PneumoniaMNIST, DermaMNIST, OCTMNIST, and OrganAMNIST. They span blood-cell microscopy, colorectal histopathology, breast ultrasound, chest radiography, dermoscopy, retinal optical coherence tomography, and abdominal CT. BreastMNIST and PneumoniaMNIST are binary; the remaining tasks contain 4 to 11 classes.

For every dataset and seed, exactly 50 images per class are sampled without replacement from the official training split. Official validation and test sets remain unchanged. Images are converted to RGB and resized to $224\times224$. Training uses batch size 16, 23 epochs, Adam with learning rate $10^{-4}$, $\epsilon=10^{-8}$, no weight decay, and no scheduler. Models are trained from scratch. The seeds are
\begin{equation}
\{3,5,7,11,13,17,19,23,29,31\}.
\end{equation}
The original \zach{} study used the first five seeds. We extend the controlled protocol to ten seeds. The PyTorch baseline is parameter-matched to the published architecture, but it is treated as the common experimental baseline rather than as an assertion of numerical replication of the earlier TensorFlow implementation.

\subsection{Controlled training conditions}
We train four conditions:
\begin{enumerate}
    \item \textbf{\zach{} + Adam}: full-precision baseline without intrinsic attribution.
    \item \textbf{\qzach{} + Adam}: W8A8 quantization-aware training and intrinsic evidence, classification loss only.
    \item \textbf{\qzach{} + Adam + $\Lattr$}: the same architecture and attribution objective as the full method, optimized by scalar loss addition.
    \item \textbf{\qzach{} + \raso{}}: the full architecture and the gradient procedure in Equations~\ref{eq:projection} and~\ref{eq:raso_gradient}.
\end{enumerate}
For each dataset-seed pair, the three \qzach{} conditions start from the same saved parameter state. Sampling manifests, initial-state hashes, checkpoints, and histories are retained. The best validation checkpoint according to the inherited primary metric is selected. All 280 expected training runs completed with no failure file.

\subsection{Actual INT8 conversion and deployment audit}
Each of the 210 selected \qzach{} checkpoints is converted independently using Equation~\ref{eq:int8_deployment}. Conversion covers seven datasets, ten seeds, and three training conditions. The official test split is then evaluated in full, yielding 964,920 matched source-to-INT8 predictions. Intrinsic deployment retention is evaluated on the original 3,600 selected XAI examples.

For every graph, we count \texttt{MatMulInteger}, \texttt{QuantizeLinear}, \texttt{DequantizeLinear}, and signed INT8 weight initializers. ONNX Runtime profiling must record integer matrix-multiplication execution. We store source and INT8 logits, probabilities, predictions, intrinsic maps, intermediate-layer parity, serialized size, and checksums. Prediction retention is summarized by exact agreement and change counts. Primary-metric retention uses Macro-F1 for multiclass datasets and AUC@0.5 for binary datasets. We additionally inspect whether a changed prediction has a source decision margin larger than twice its corresponding maximum logit perturbation; this is reported only as a numerical audit, not as a standard robustness metric.

Batch-size-1 CPU latency is measured after ten warm-up iterations over 50 timed iterations, with one and four threads. The source path is PyTorch 2.3.1 eager fake-QAT execution; the deployed path is ONNX Runtime 1.23.2 with \texttt{CPUExecutionProvider}. Consequently, the latency ratio is an end-to-end deployment comparison that includes both quantization and runtime-stack differences, not a pure isolated-kernel speedup. The run used Python 3.10.16, ONNX 1.22.0, MedMNIST 3.0.2, and a 32-logical-CPU x86\_64 Linux host. The exact processor model was not recorded by the audit script.

\subsection{Predictive metrics and statistical analysis}
Macro-F1 is primary for multiclass datasets. For continuity with the predecessor protocol, binary primary performance is the ROC-AUC of predictions thresholded at 0.5, denoted AUC@0.5. Because this is not conventional probability AUROC, probability AUROC, accuracy, and Macro-F1 are also retained and reported in Appendix~\ref{app:binary}. The main \qzach{} predictive table and all paired contrasts use the actual ONNX INT8 outputs. The \zach{} baseline remains FP32.

Results are mean $\pm$ standard deviation across ten seeds. Dataset-level mean ranks are compared with a Friedman test across the seven datasets. Within each dataset, paired two-sided Wilcoxon signed-rank tests compare conditions over ten seeds, with Holm correction across seven datasets for each planned contrast. The 70 dataset-seed differences are also reported descriptively through their mean and number of positive pairs.

\subsection{XAI baselines and compatibility audit}
The full same-model comparison explains the source-model predictions of each trained \qzach{} + \raso{} checkpoint with:
\begin{itemize}
    \item the intrinsic recursive map;
    \item Integrated Gradients with 32 integration steps;
    \item token-adapted Grad-CAM;
    \item Attention Rollout-ZT;
    \item Gradient Attention Rollout-ZT;
    \item RISE with 128 masks;
    \item deterministic patch occlusion;
    \item random attribution as a negative control.
\end{itemize}
The gradient-based post-hoc comparison is performed on the differentiable PyTorch source model. The actual ONNX INT8 deployment is evaluated separately for intrinsic-map retention because the integer graph is not used as a differentiable post-hoc explanation backend. ZACH-ViT has no classification token. The two rollout methods therefore average final-query contribution in a manner consistent with global average pooling rather than extracting a nonexistent classification-token row. Exact transformer LRP is excluded because no relevance-conservation rule was derived for the custom quantized attention, adaptive residual projections, recursive evidence fusion, and zero-token aggregation.

The benchmark selects two examples per class where possible, with 16 to 32 examples per dataset-seed checkpoint. This yields 1,200 matched explanations per method for the \qzach{} + \raso{} same-model comparison, 1,200 intrinsic explanations for each of the three \qzach{} conditions, and 1,200 explanations per post-hoc method on the \zach{} baseline. In total, 21,600 source-model XAI rows are evaluated, plus 168 parameter-randomization rows and 3,600 source-to-INT8 intrinsic-map pairs.

\subsection{Faithfulness, stability, and sanity metrics}
Faithfulness is assessed by deletion AUC, insertion AUC, comprehensiveness after removing the top 20\% patches, sufficiency error when retaining those patches, SaCo with ten salience groups, and rank correlation with deterministic patch occlusion. Stability under small input noise uses Gaussian noise with standard deviation 0.01 and reports cosine similarity, rank correlation, top-10\% overlap, and JS divergence. Runtime per image, approximate forward passes, and approximate backward passes are recorded.

The official MedMNIST NPZ archives were inspected directly. They contain split images and classification labels but no masks, bounding boxes, landmarks, or localization targets. We therefore do not compute Dice, intersection-over-union, or pointing-game metrics. Parameter-randomization checks compare original and fully randomized models on four examples per dataset.

For intrinsic-XAI comparisons, each metric is first averaged over matched examples within a dataset-seed checkpoint, yielding 70 paired units per condition. \raso{} is compared with the attribution-loss control by two-sided Wilcoxon tests, with Holm correction across ten XAI metrics. We emphasize effect direction and multiplicity-corrected results rather than isolated uncorrected $p$-values.

\section{Results}
\subsection{Actual INT8 predictive performance}
Table~\ref{tab:main_prediction} reports the FP32 \zach{} baseline and the actual converted ONNX INT8 outputs for all three \qzach{} conditions. Every deployed \qzach{} condition exceeds the \zach{} mean on every dataset. Relative to the baseline across 70 dataset-seed pairs, actual INT8 \qzach{} + Adam has a mean gain of 0.0313 and improves 53 pairs. The attribution-loss control has a mean gain of 0.0309 and improves 52 pairs. Actual INT8 \qzach{} + \raso{} has the largest mean gain, 0.0368, and improves 57 pairs.

\begin{table}[t]
\centering
\caption{Test performance under the 50-images-per-class protocol. The \zach{} baseline is FP32; all \qzach{} values are from the actual converted ONNX INT8 graphs. Values are mean $\pm$ standard deviation over ten seeds. Macro-F1 is used for multiclass datasets and AUC@0.5 for binary datasets. The best mean in each row is bold.}
\label{tab:main_prediction}
\scriptsize
\resizebox{\textwidth}{!}{%
\begin{tabular}{llcccc}
\toprule
Dataset & Metric & \zach{} + Adam & \qzach{} + Adam & \qzach{} + Adam + $\Lattr$ & \qzach{} + \raso{} \\
\midrule
BloodMNIST & Macro-F1 & 0.675 $\pm$ 0.039 & 0.723 $\pm$ 0.026 & 0.722 $\pm$ 0.018 & \textbf{0.733 $\pm$ 0.024} \\
PathMNIST & Macro-F1 & 0.504 $\pm$ 0.059 & 0.571 $\pm$ 0.026 & \textbf{0.577 $\pm$ 0.029} & 0.564 $\pm$ 0.031 \\
BreastMNIST & AUC@0.5 & 0.584 $\pm$ 0.045 & 0.619 $\pm$ 0.051 & 0.613 $\pm$ 0.053 & \textbf{0.620 $\pm$ 0.045} \\
PneumoniaMNIST & AUC@0.5 & 0.738 $\pm$ 0.013 & 0.746 $\pm$ 0.016 & \textbf{0.748 $\pm$ 0.018} & 0.745 $\pm$ 0.021 \\
DermaMNIST & Macro-F1 & 0.340 $\pm$ 0.021 & \textbf{0.354 $\pm$ 0.021} & 0.348 $\pm$ 0.023 & 0.345 $\pm$ 0.015 \\
OCTMNIST & Macro-F1 & 0.238 $\pm$ 0.025 & 0.262 $\pm$ 0.034 & 0.263 $\pm$ 0.035 & \textbf{0.274 $\pm$ 0.026} \\
OrganAMNIST & Macro-F1 & 0.463 $\pm$ 0.022 & 0.487 $\pm$ 0.035 & 0.490 $\pm$ 0.031 & \textbf{0.520 $\pm$ 0.030} \\
\bottomrule
\end{tabular}}
\end{table}

The best deployed \qzach{} condition remains task dependent. \raso{} has the best mean on BloodMNIST, BreastMNIST, OCTMNIST, and OrganAMNIST; the attribution-loss control is best on PathMNIST and PneumoniaMNIST; and Adam is best on DermaMNIST. Mean ranks are 1.86 for \raso{}, 2.00 for the attribution-loss control, 2.14 for Adam, and 4.00 for \zach{}. The Friedman test rejects equal ranks ($\chi^2_F=12.77$, $p=0.0052$), but the small and heterogeneous differences among the three deployed \qzach{} conditions do not support a universal optimizer ranking.

Across the seven dataset means, each deployed \qzach{} condition is higher than \zach{} on all seven tasks, giving a two-sided signed-rank value of $p=0.0156$ for each comparison. After per-contrast Holm correction across datasets, actual INT8 \qzach{} + Adam is significant on BloodMNIST and PathMNIST. Actual INT8 \qzach{} + \raso{} is significant on BloodMNIST, PathMNIST, OCTMNIST, and OrganAMNIST. Full paired results are given in Appendix~\ref{app:stats}.

\begin{figure}[t]
\centering
\includegraphics[width=0.94\textwidth]{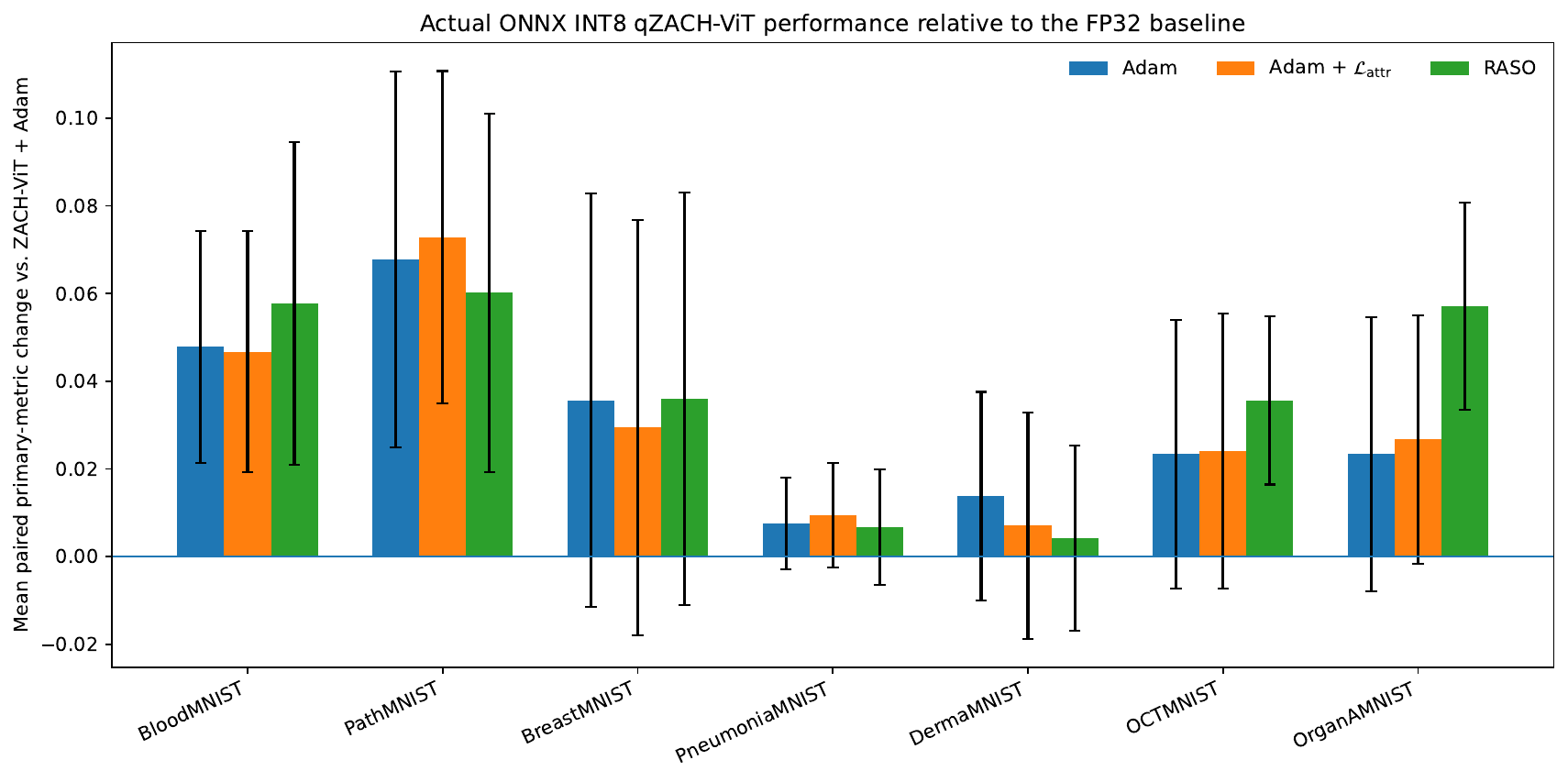}
\caption{Mean paired change of the actual ONNX INT8 \qzach{} models relative to the FP32 \zach{} baseline. Error bars are 95\% confidence intervals across the ten seed-wise differences within each dataset.}
\label{fig:prediction_delta}
\end{figure}

The architectural result therefore survives real conversion: quantization-aware intrinsic patch evidence does not impose an observed predictive penalty in this protocol, and every deployed \qzach{} dataset mean remains above the FP32 baseline. \raso{} adds an average 0.0059 over the deployed loss-only control across 70 pairs and is positive in 42 pairs, but this additional predictive effect is heterogeneous.

\subsection{Exact INT8 deployment fidelity}
All 210 selected \qzach{} checkpoints were converted successfully. Every ONNX graph contains 16 signed INT8 \texttt{MatMulInteger} projections and 16 signed INT8 weight initializers, and every ONNX Runtime profile records integer matrix-multiplication execution. Across 964,920 matched test predictions, 240 labels change after conversion, corresponding to 99.9751\% exact agreement. The mean absolute change in the inherited primary metric is 0.000133 across checkpoints, the maximum is 0.004386, and 126 of 210 checkpoints have an exactly unchanged inherited primary metric.

\begin{table}[t]
\centering
\caption{Actual ONNX INT8 retention by training condition. Prediction counts aggregate all official test images. Primary-metric changes are absolute checkpoint-level changes relative to the corresponding source QAT model. Intrinsic-XAI values aggregate the matched deployment maps.}
\label{tab:int8_retention}
\scriptsize
\resizebox{\textwidth}{!}{%
\begin{tabular}{lrrrrrrr}
\toprule
Condition & Changed / total & Agreement & Mean $|\Delta|$ & Max $|\Delta|$ & Map cosine & Rank $\rho$ & Top-10\% overlap \\
\midrule
\qzach{} + Adam & 95/321,640 & 99.9705\% & 0.000124 & 0.004386 & 0.999948 & 0.993851 & 0.9691 \\
\qzach{} + Adam + $\Lattr$ & 89/321,640 & 99.9723\% & 0.000166 & 0.004386 & 0.999948 & 0.994655 & 0.9701 \\
\qzach{} + \raso{} & 56/321,640 & 99.9826\% & 0.000110 & 0.001282 & 0.999968 & 0.994586 & 0.9684 \\
\bottomrule
\end{tabular}}
\end{table}

The smallest number of changed predictions occurs for \raso{}: 56 of 321,640, compared with 95 for Adam and 89 for the loss-only control. No changed prediction has a source decision margin larger than twice its corresponding maximum logit perturbation. This margin diagnostic is not a general robustness guarantee, but it shows that all 240 label changes occur near a source decision boundary relative to the observed conversion error. Across all samples, the median per-sample maximum logit error is 0.000203, the 99th percentile is 0.0145, and the maximum is 0.1585; the isolated maximum does not translate into a large primary-metric change.

The deployment also preserves intrinsic evidence. Across 3,600 source-to-INT8 map pairs, mean recursive cosine is 0.999955, mean JS divergence is $1.13\times10^{-5}$, mean Spearman patch-rank correlation is 0.9944, and mean top-10\% overlap is 0.9692. The worst individual recursive cosine is 0.997916. \raso{} has the highest mean recursive cosine, 0.999969, and the lowest mean recursive JS divergence among the three conditions, but it does not have the highest top-patch overlap. These results support strong deployment retention without claiming bitwise identity.

\begin{figure}[t]
\centering
\includegraphics[width=0.82\textwidth]{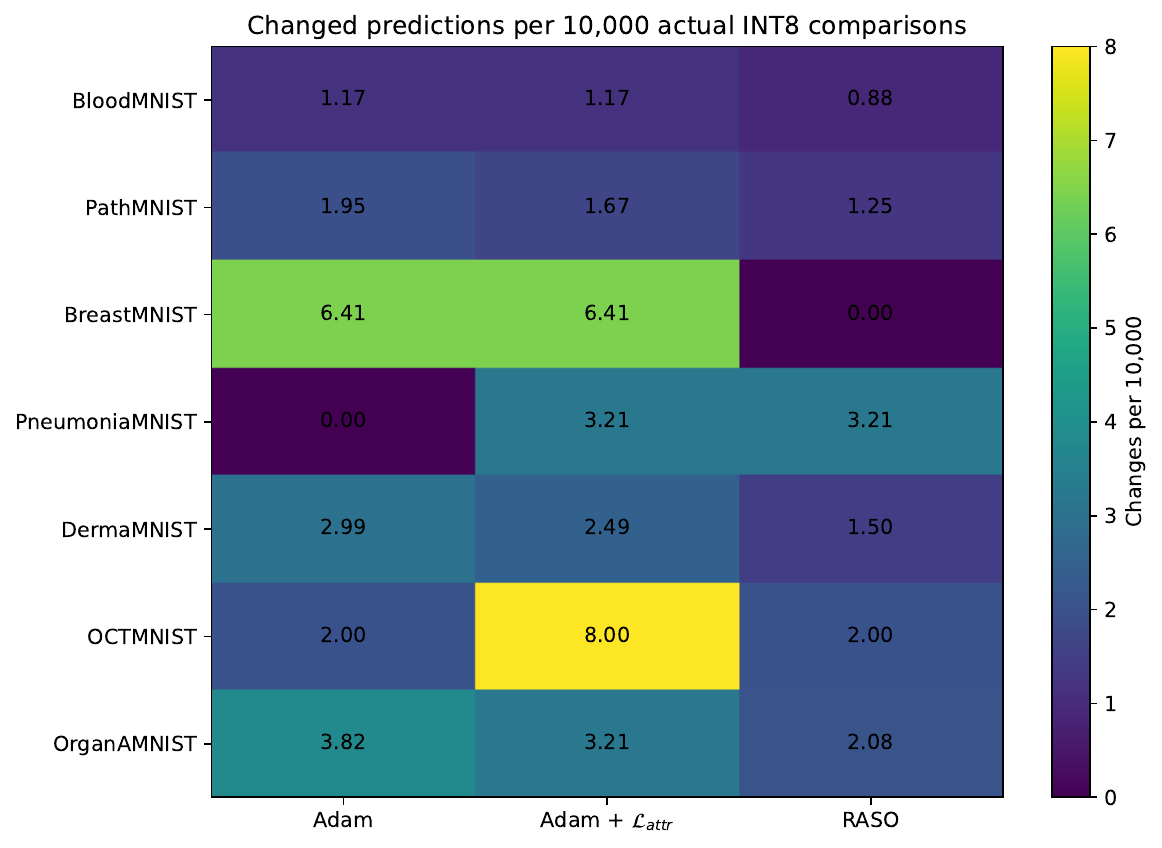}
\caption{Changed predictions per 10,000 source-to-INT8 comparisons. Values use all official test images for every dataset, condition, and seed.}
\label{fig:int8_changes}
\end{figure}

\subsection{Serialized size and CPU latency}
The mean serialized source checkpoint is 0.988 MiB and the mean exact INT8 ONNX artifact is 0.296 MiB, a 70.03\% reduction. This is measured file size, including graph and container overhead, rather than theoretical parameter accounting. Dataset-specific values are reported in Appendix~\ref{app:int8_details}.

\begin{table}[t]
\centering
\caption{Batch-size-1 CPU latency. Values are the mean $\pm$ standard deviation of checkpoint-level median latency over 210 models. Speedup is computed per checkpoint and then averaged.}
\label{tab:int8_latency}
\scriptsize
\begin{tabular}{rcccc}
\toprule
Threads & Source PyTorch QAT (ms) & Actual INT8 ONNX (ms) & Mean speedup & Checkpoint range \\
\midrule
1 & 6.853 $\pm$ 0.074 & 4.876 $\pm$ 0.038 & 1.41$\times$ & 1.38 to 1.50$\times$ \\
4 & 5.870 $\pm$ 0.131 & 2.456 $\pm$ 0.083 & 2.39$\times$ & 2.05 to 2.57$\times$ \\
\bottomrule
\end{tabular}
\end{table}

At one thread, median latency averaged across checkpoints decreases from 6.853 ms to 4.876 ms, a mean 1.41 times speedup. At four threads, it decreases from 5.870 ms to 2.456 ms, a mean 2.39 times speedup. These are end-to-end comparisons between PyTorch eager fake-QAT execution and ONNX Runtime exact INT8 execution. They should not be interpreted as an isolated quantization-kernel ablation.

\begin{figure}[t]
\centering
\includegraphics[width=0.82\textwidth]{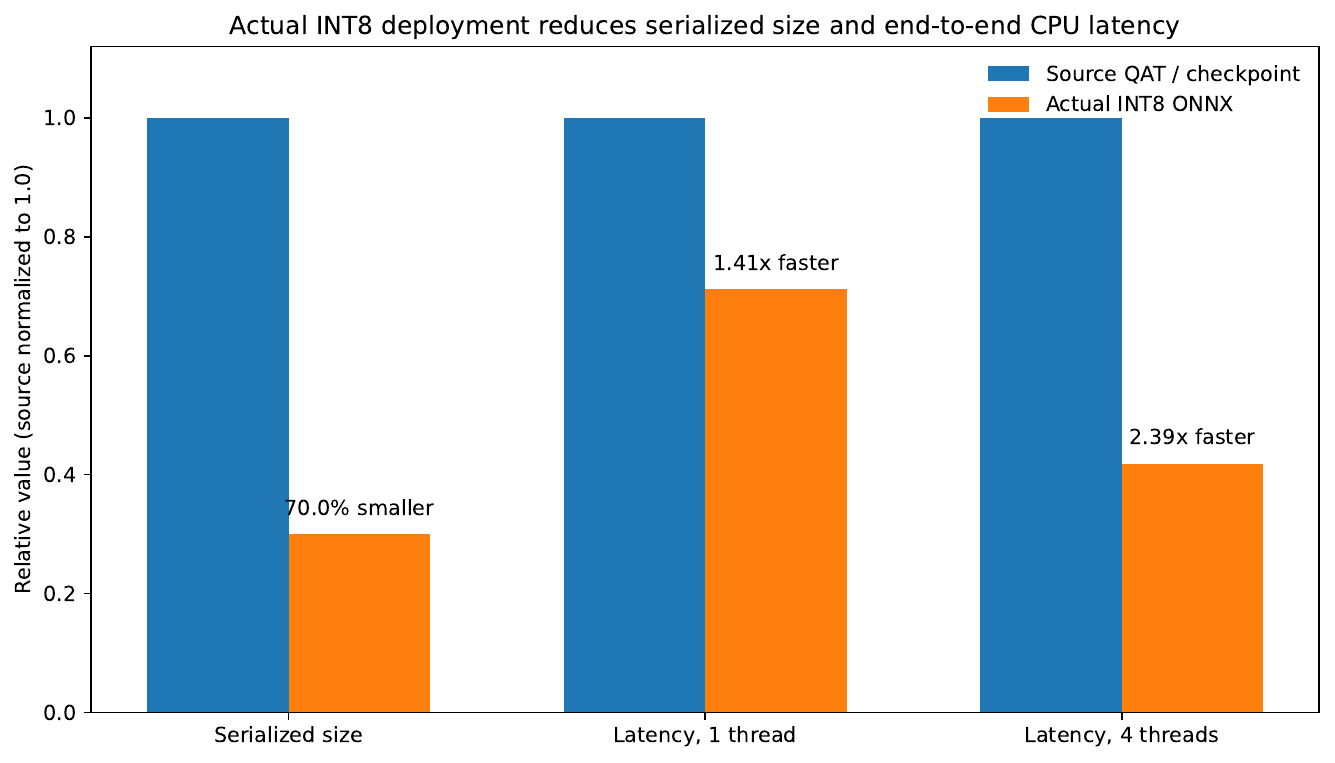}
\caption{Serialized size and CPU latency normalized independently to the corresponding source value. The actual INT8 ONNX artifact is 70.0\% smaller, and end-to-end inference is 1.41 times and 2.39 times faster with one and four threads, respectively.}
\label{fig:int8_efficiency}
\end{figure}

\subsection{What RASO changes in intrinsic XAI}
Table~\ref{tab:intrinsic_ablation} compares the intrinsic explanation across the three differentiable source \qzach{} conditions. \raso{} has the lowest deletion AUC and sufficiency error, and the highest noise cosine. It also has the lowest noise JS divergence. It is not best on insertion AUC, comprehensiveness, SaCo, occlusion correlation, or every top-patch stability measure.

\begin{table}[t]
\centering
\caption{Source-model intrinsic-XAI comparison. Each value averages 1,200 matched explanations. Del. is deletion AUC, Ins. is insertion AUC, Comp. is comprehensiveness, Suff. is sufficiency error, Occ. $\rho$ is correlation with patch occlusion, and Noise JS is divergence under input noise.}
\label{tab:intrinsic_ablation}
\scriptsize
\resizebox{\textwidth}{!}{%
\begin{tabular}{lcccccccc}
\toprule
Condition & Del. $\downarrow$ & Ins. $\uparrow$ & Comp. $\uparrow$ & Suff. $\downarrow$ & SaCo $\uparrow$ & Occ. $\rho$ $\uparrow$ & Noise cos. $\uparrow$ & Noise JS $\downarrow$ \\
\midrule
qZACH-ViT + Adam & 0.352 & 0.375 & 0.190 & 0.339 & 0.205 & 0.116 & 0.991 & 0.003 \\
qZACH-ViT + Adam + $\mathcal{L}_{\mathrm{attr}}$ & 0.344 & 0.367 & 0.186 & 0.335 & 0.217 & 0.088 & 0.993 & 0.002 \\
qZACH-ViT + RASO & 0.339 & 0.364 & 0.175 & 0.304 & 0.207 & 0.108 & 0.995 & 0.002 \\
\bottomrule
\end{tabular}}
\end{table}

After aggregating examples within each dataset-seed checkpoint and applying Holm correction across ten metrics, \raso{} improves three measures significantly relative to Adam with the same attribution loss: sufficiency error decreases by 0.0300 ($p_{\mathrm{Holm}}=0.0111$), noise cosine increases by 0.00177 ($p_{\mathrm{Holm}}<0.0001$), and noise JS decreases by 0.00055 ($p_{\mathrm{Holm}}<0.0001$). The other seven metrics do not survive correction. This supports a targeted stability and sufficiency claim, not broad superiority across all notions of faithfulness.

\begin{figure}[t]
\centering
\includegraphics[width=0.88\textwidth]{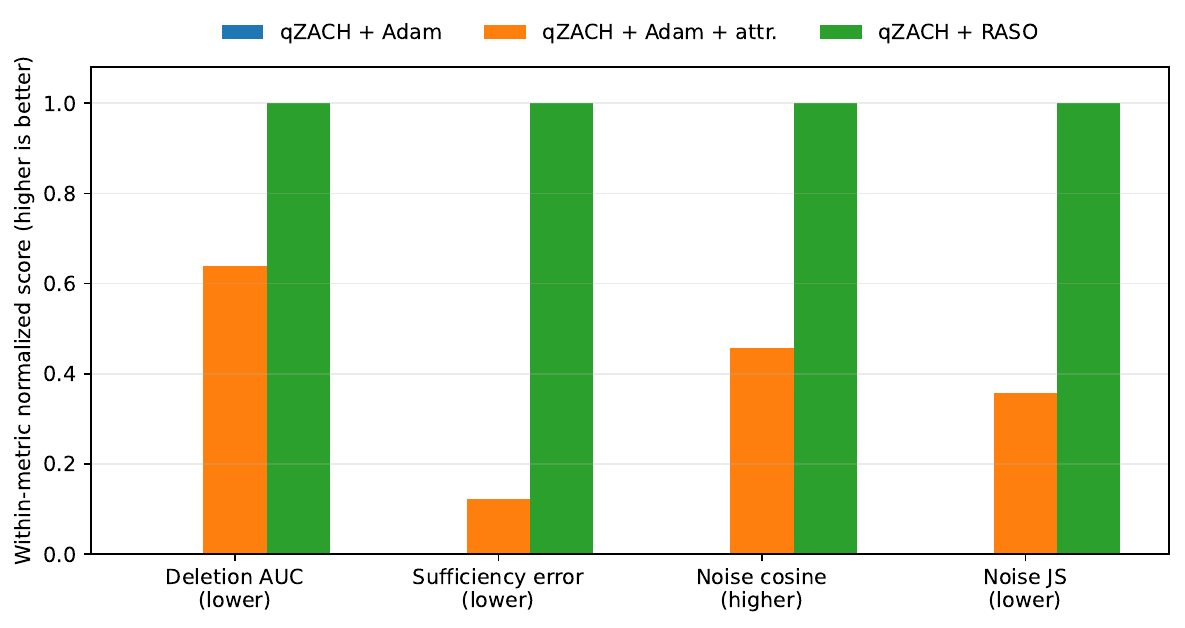}
\caption{Selected source-model intrinsic-XAI metrics normalized within each metric for visualization. Raw values and statistical tests are reported in Table~\ref{tab:intrinsic_ablation} and Appendix~\ref{app:xai_stats}.}
\label{fig:intrinsic_ablation}
\end{figure}

The projection mechanism is active rather than dormant. Classification and attribution gradients conflict in 45.9\% of training batches on average. Dataset means range from 37.7\% on OCTMNIST to 53.4\% on OrganAMNIST.

\begin{figure}[t]
\centering
\includegraphics[width=0.82\textwidth]{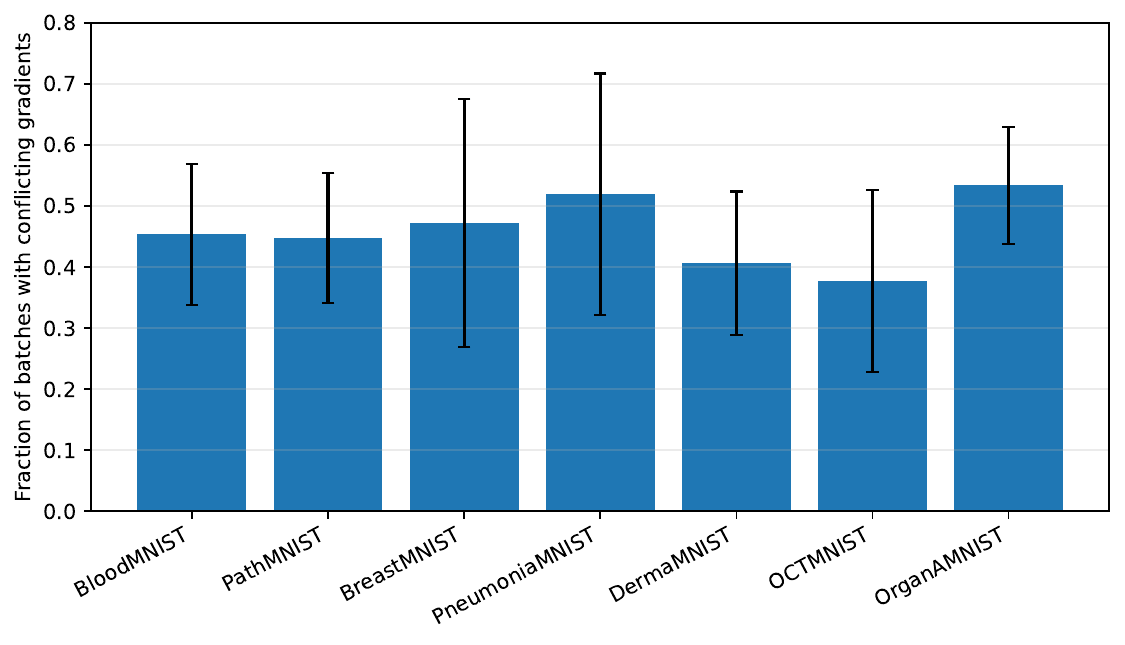}
\caption{Mean fraction of \raso{} training batches with a negative classification-attribution gradient inner product. Error bars show variation across epochs and seeds.}
\label{fig:gradient_conflict}
\end{figure}

\subsection{Same-model comparison with post-hoc methods}
Table~\ref{tab:same_model_xai} compares explanations for exactly the same source \qzach{} + \raso{} predictions. Gradient Attention Rollout-ZT is the strongest non-oracle method on deletion, insertion, SaCo, and correlation with patch occlusion. The intrinsic method is therefore not presented as uniformly state of the art in post-hoc faithfulness. It is competitive with Token Grad-CAM on deletion and occlusion correlation, exceeds RISE on deletion and SaCo, and exceeds random attribution on deletion, comprehensiveness, sufficiency, SaCo, and occlusion correlation. Random attribution is slightly higher on insertion AUC, illustrating why no single perturbation metric is sufficient. The actual INT8 retention results above show that the intrinsic maps evaluated here are preserved after deployment conversion.

\begin{table}[t]
\centering
\caption{Same-model source-XAI comparison on \qzach{} + \raso{}. Each row summarizes 1,200 matched explanations. Patch occlusion is both an explainer and the reference for Occ. $\rho$, so its value of 1 is tautological.}
\label{tab:same_model_xai}
\scriptsize
\begin{tabular}{lcccccc}
\toprule
Method & Del. $\downarrow$ & Ins. $\uparrow$ & SaCo $\uparrow$ & Occ. $\rho$ $\uparrow$ & Noise cos. $\uparrow$ & ms/image $\downarrow$ \\
\midrule
\textbf{qZACH intrinsic} & 0.339 & 0.364 & 0.207 & 0.108 & 0.995 & 9.9 \\
Integrated Gradients & 0.310 & 0.388 & 0.220 & 0.036 & 0.990 & 623.4 \\
Token Grad-CAM & 0.341 & 0.364 & 0.188 & 0.109 & 0.989 & 21.1 \\
Attention Rollout-ZT & 0.302 & 0.384 & 0.304 & 0.047 & 0.994 & 11.3 \\
Gradient Attention Rollout-ZT & 0.279 & 0.423 & 0.470 & 0.196 & 1.000 & 22.4 \\
RISE & 0.356 & 0.385 & 0.089 & 0.011 & 1.000 & 45.6 \\
Patch Occlusion & 0.302 & 0.409 & 0.602 & 1.000 & 0.908 & 81.4 \\
Random Attribution & 0.376 & 0.376 & 0.005 & 0.001 & 1.000 & 0.4 \\
\bottomrule
\end{tabular}
\end{table}

The intrinsic map requires one forward pass and no backward pass, averaging 9.9 ms per image in the matched source-model XAI implementation. Integrated Gradients averages 623.4 ms, Token Grad-CAM 21.1 ms, Gradient Attention Rollout-ZT 22.4 ms, RISE 45.6 ms, and patch occlusion 81.4 ms. Attention Rollout-ZT is the closest at 11.3 ms. These XAI runtimes are separate from the logits-only deployment latency in Table~\ref{tab:int8_latency}.

\begin{figure}[t]
\centering
\includegraphics[width=0.82\textwidth]{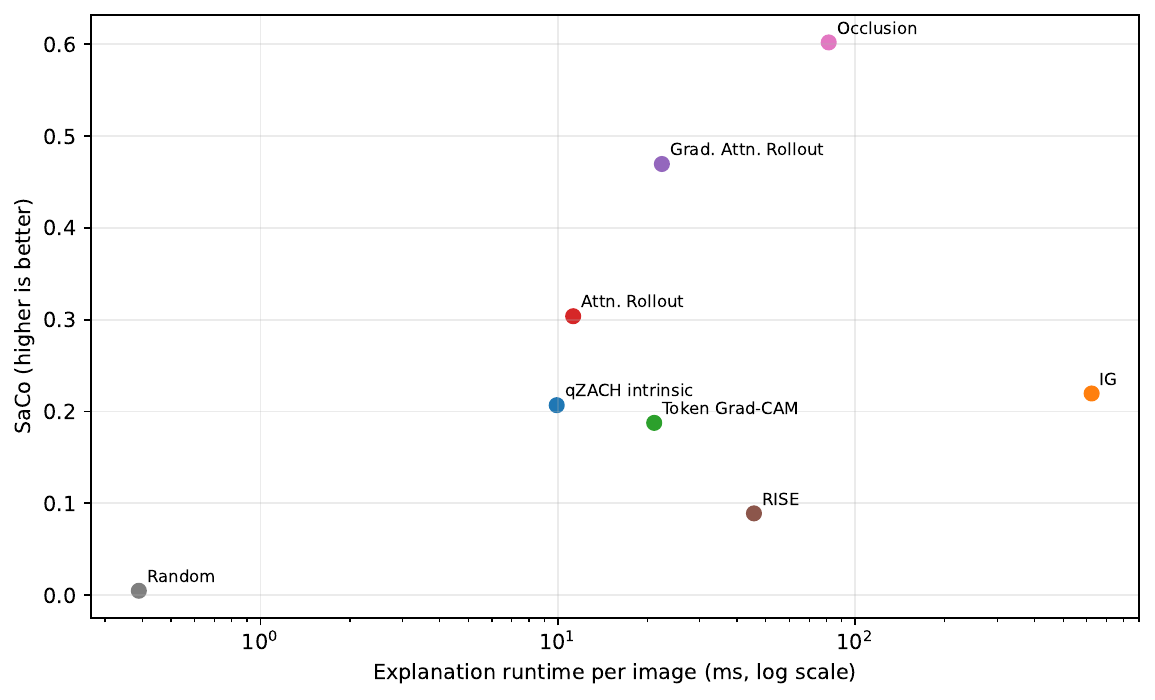}
\caption{SaCo versus explanation runtime for the same source \qzach{} + \raso{} predictions. The intrinsic explanation lies on the fast end of the tradeoff but is not the most faithful by SaCo.}
\label{fig:xai_runtime}
\end{figure}

\subsection{Parameter-randomization sanity checks}
A faithful explainer should depend on learned parameters. Integrated Gradients changes most strongly after full parameter randomization, with cosine 0.628 and JS 0.089. The intrinsic map retains a high raw cosine of 0.963, but its patch-rank correlation becomes negative and its top-10\% overlap falls to 0.109, close to the 0.10 chance level. This combination shows why cosine alone is unreliable for smooth nonnegative maps. Gradient Attention Rollout-ZT and RISE have near-unit cosine and extremely small JS after randomization, although their rank and top-patch overlap also deteriorate. We therefore interpret their strong perturbation metrics cautiously. Full values are in Appendix~\ref{app:sanity}.

\subsection{Qualitative behavior}
\begin{figure}[p]
\centering
\includegraphics[width=\textwidth]{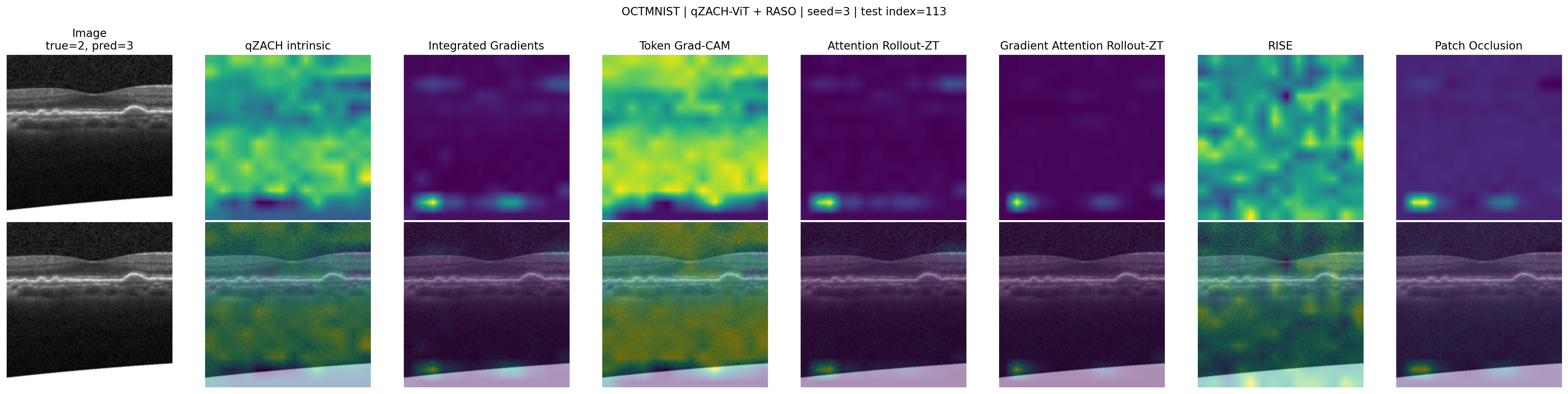}
\caption{Matched XAI methods for one OCTMNIST test image and one source \qzach{} + \raso{} checkpoint. The original and overlay are shown for each method. The model predicts class 3 while the true class is 2.}
\label{fig:qualitative_same}
\end{figure}

\begin{figure}[p]
\centering
\includegraphics[width=\textwidth]{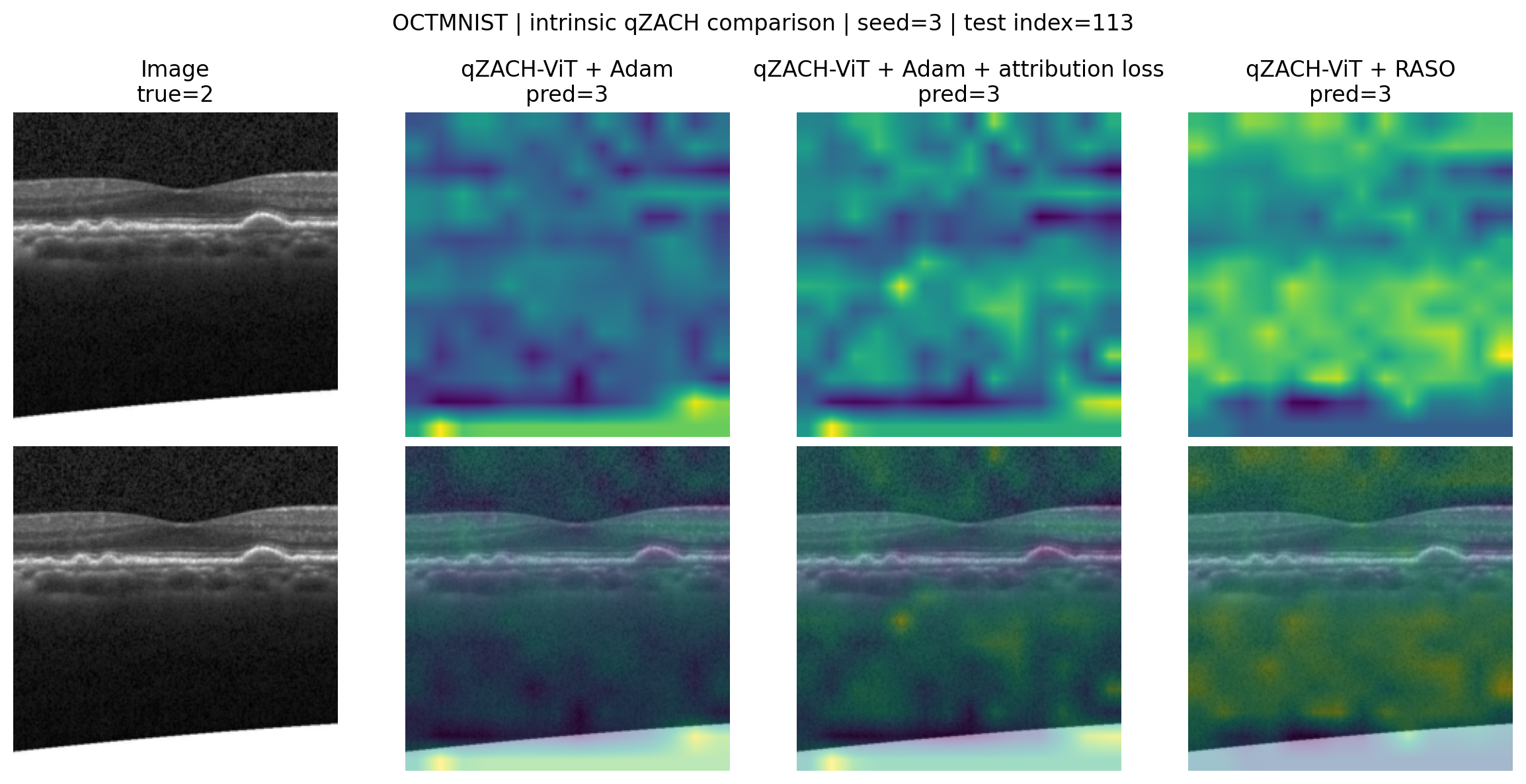}
\caption{Intrinsic maps from the three source \qzach{} conditions for the same OCTMNIST image. The visual difference is descriptive; no anatomical localization target is available.}
\label{fig:qualitative_three}
\end{figure}

Figure~\ref{fig:qualitative_same} shows a matched OCTMNIST example. The explainers differ markedly despite sharing the same prediction. Gradient and occlusion methods emphasize a narrow lower region, while the intrinsic and token Grad-CAM maps are more diffuse. Figure~\ref{fig:qualitative_three} compares the three intrinsic conditions. \raso{} produces a smoother recursive allocation in this example, but the prediction is incorrect and no localization mask exists. The figure must therefore be read as a visualization of model evidence, not proof of medical correctness.

\section{Discussion}
The results support three separate conclusions.

First, \qzach{} is a viable architectural extension of the compact backbone, and this conclusion survives actual deployment. The classification-only \qzach{} condition improves 53 of 70 paired runs over the FP32 baseline before an attribution objective is introduced. After conversion, every \qzach{} dataset mean remains above the \zach{} mean, and the mean absolute checkpoint-level primary-metric change from source QAT to actual INT8 is only 0.000133. Intrinsic explanation is therefore not purchased at the cost of an observed predictive collapse in this protocol.

Second, the deployment evidence is stronger than a fake-quantization result. All 210 ONNX models contain and execute 16 signed INT8 projection operators. Across nearly one million predictions, exact agreement is 99.9751\%, while 3,600 intrinsic maps retain mean cosine 0.999955, mean rank correlation 0.9944, and mean top-10\% overlap 0.9692. The actual artifacts are 70.0\% smaller than the serialized source checkpoints and faster in the measured end-to-end CPU setting. These findings show that both decisions and intrinsic evidence survive conversion. They do not establish fully integer-only execution because several non-projection operations remain FP32, and the latency comparison includes a change of runtime stack.

Third, \raso{} changes explanation behavior in a narrower and more defensible way than a universal optimizer claim would suggest. It obtains the largest mean predictive gain over the baseline, the fewest source-to-INT8 changed predictions, and the highest mean deployment-map cosine. In the source-model XAI benchmark, it significantly improves sufficiency error and input-noise stability over Adam with the identical attribution loss. However, it is not the best condition on every dataset, does not lead every faithfulness metric, and has slightly lower mean top-10\% deployment overlap than the other two \qzach{} conditions. Its contribution is best described as stabilizing selected properties of recursive intrinsic evidence while protecting classification.

The comparison with post-hoc methods further clarifies the tradeoff. The intrinsic map is available without a backward pass or repeated perturbation, making it substantially faster than Integrated Gradients, RISE, and occlusion in the matched source implementation. Gradient Attention Rollout-ZT nevertheless yields stronger average deletion, insertion, SaCo, and occlusion-correlation values. Intrinsic availability, deployability, and computational efficiency therefore do not automatically imply maximal perturbation faithfulness.

The architecture also exposes a conceptual advantage. Raw patch evidence is not an auxiliary saliency head trained only to imitate a post-hoc map. It is the quantity whose spatial mean produces the stage logit. This exact completeness is stronger than claiming that an attention matrix explains a prediction. The normalized heatmap remains a transformation of that evidence, and its clinical relevance is unverified, but the evidence path is part of the predictor and remains numerically stable after INT8 conversion.

\raso{} should be understood in relation to gradient-surgery methods. Its projection resembles the normal-plane operation used in PCGrad, while norm matching resembles the motivation of GradNorm. The novelty is the asymmetric combination and its use for a recursive quantization-consistency objective with a classification-protected direction. Direct comparisons with PCGrad, CAGrad, GradNorm, and MGDA remain important for a later conference version. The present loss-only control establishes that the gradient procedure can add value beyond scalar regularization, but it is not a complete optimizer benchmark.

Finally, this study demonstrates the value of multiple XAI criteria and multiple quantization checks. Smooth maps can retain high cosine after model randomization, and a single maximum numerical error can look large without changing decisions or task metrics. The analysis therefore combines graph inspection, runtime profiling, full-test prediction agreement, primary-metric retention, rank-based map retention, perturbation tests, stability, and randomization rather than relying on one favorable scalar.

\section{Limitations}
\begin{enumerate}
    \item \textbf{Mixed precision rather than fully integer-only inference.} All 16 learned projections use signed INT8 \texttt{MatMulInteger} with INT32 accumulation, but LayerNorm, residual additions, ReLU, attention score and context products, softmax, pooling, dequantization, bias addition, and attribution normalization remain FP32.
    \item \textbf{Runtime-stack confounding.} The latency comparison is between PyTorch eager fake-QAT execution and ONNX Runtime actual INT8 execution. It is a valid end-to-end deployment comparison for the tested software stack, but it does not isolate quantization from graph optimization and runtime implementation.
    \item \textbf{Incomplete hardware metadata.} The audit records an x86\_64 Linux host with 32 logical CPUs and all software versions, but the exact processor model was not captured. Latency values should therefore not be treated as universal hardware benchmarks.
    \item \textbf{No peak-memory claim.} Serialized artifact size is measured directly, but reliable process-level peak-memory reduction was not established and is not claimed.
    \item \textbf{No ground-truth localization.} The seven MedMNIST archives contain classification labels only. Faithfulness is evaluated through perturbation, sufficiency, stability, deployment retention, and randomization rather than anatomical masks or expert regions.
    \item \textbf{Single benchmark family.} The seven tasks cover several modalities, but all are distributed through MedMNIST and share standardized preprocessing. External blood microscopy, histopathology, dermoscopy, and XAI-specific datasets are needed to assess transfer.
    \item \textbf{Coarse patch resolution.} A $14\times14$ grid limits spatial precision. Upsampled heatmaps may appear more anatomically precise than the underlying patch evidence.
    \item \textbf{RASO baselines.} The optimizer-specific control is Adam with the identical attribution loss. Direct comparisons with PCGrad, CAGrad, GradNorm, and other multi-objective methods are deferred.
    \item \textbf{Framework port.} The full-precision baseline is a parameter-matched PyTorch implementation of the published architecture. It is not claimed to reproduce every earlier TensorFlow result numerically. Internal comparisons remain controlled because all four conditions share the same framework, samples, and training code.
    \item \textbf{Metric dependence.} Deletion and insertion can introduce out-of-distribution inputs. SaCo, sufficiency, randomization, and the margin-based conversion audit also have assumptions. No single score is treated as ground truth.
    \item \textbf{No clinical claim.} The heatmaps indicate model evidence, not verified pathology, diagnosis, or causal anatomy. Expert review and task-specific localization data are required before clinical interpretation.
\end{enumerate}

\section{Conclusion}
We introduced \qzach{} and \raso{} as two distinct contributions and validated them beyond quantization simulation. \qzach{} combines a compact zero-token Vision Transformer with W8A8 quantization-aware training and recursive patch-level evidence that exactly reconstructs the logits before normalization. The selected models were converted into 210 executable mixed-precision ONNX INT8 graphs in which all 16 learned projections use signed INT8 \texttt{MatMulInteger} with INT32 accumulation. Across seven low-data MedMNIST tasks and ten seeds, the actual deployed \qzach{} means exceed the FP32 \zach{} baseline on every dataset. Across 964,920 source-to-INT8 test comparisons, prediction agreement is 99.9751\%, and across 3,600 intrinsic-map pairs, mean cosine is 0.999955. The actual ONNX artifacts are 70.0\% smaller than the serialized source checkpoints and provide 1.41 times and 2.39 times end-to-end CPU speedups with one and four threads in the measured environment. \raso{} yields the largest overall predictive gain, the fewest changed deployment predictions, and targeted improvements in sufficiency and input-noise stability, but it is not universally best across datasets or XAI metrics. Strong post-hoc methods remain superior on several faithfulness criteria. These balanced results establish \qzach{} as a genuinely deployable compact intrinsically explainable model and \raso{} as a stability-oriented optimization procedure. External datasets, localization-grounded evaluation, fully integer non-projection operators, complete hardware profiling, and broader optimizer comparisons remain the principal next steps.

\section*{Reproducibility Statement}
The training archive contains the configuration, method lock, 280 checkpoints, 70 sampling manifests, locked initial states for the three \qzach{} conditions, complete training histories, seed-level and summary tables, the XAI compatibility audit, 21,600 source-model XAI rows, 168 parameter-randomization rows, qualitative panels, and a completion audit. The deployment archive contains 210 ONNX models, 210 complete prediction archives, 210 intrinsic-XAI retention files, 210 graph audits, 210 ONNX Runtime execution audits, 1,260 intermediate-layer parity rows, 840 latency measurements, checksums, and zero failure files. The manuscript source additionally includes recomputed deployment summaries and a corrected audit derived from the saved prediction and XAI files. Exact seeds, hyperparameters, quantization rules, operator coverage, software versions, and evaluation settings are stated in the manuscript.

\appendix
\section{Detailed Architecture and Hardware-Deployed INT8}
\subsection{Architecture and actual serialized deployment size}
All models use 196 patch tokens. The architecture audit verifies that the recursive raw patch evidence reconstructs the logits with zero numerical error at floating-point precision for every dataset-specific output dimension. Quantized weights include the patch projection, all attention projections, feed-forward projections, adaptive residual projections, token MLP projections, and evidence-head weights.

Table~\ref{tab:int8_storage} reports measured serialized file sizes rather than theoretical parameter accounting. Source size is the mean selected PyTorch QAT checkpoint size across the three \qzach{} conditions and ten seeds. ONNX size is the mean actual converted artifact size. Each ONNX graph contains 16 \texttt{MatMulInteger} projection nodes.

\begin{table}[H]
\centering
\caption{Architecture and actual serialized deployment size. Sizes are MiB. Reduction is relative to the corresponding serialized source checkpoint.}
\label{tab:int8_storage}
\scriptsize
\begin{tabular}{lrrrrrr}
\toprule
Dataset & qZACH params & INT8 weights & Source checkpoint & INT8 ONNX & Reduction & Integer projections \\
\midrule
BloodMNIST & 249,744 & 247,296 & 0.990 & 0.297 & 70.04\% & 16 \\
PathMNIST & 249,938 & 247,488 & 0.990 & 0.297 & 70.04\% & 16 \\
BreastMNIST & 248,386 & 245,952 & 0.985 & 0.295 & 70.02\% & 16 \\
PneumoniaMNIST & 248,386 & 245,952 & 0.985 & 0.295 & 70.03\% & 16 \\
DermaMNIST & 249,550 & 247,104 & 0.989 & 0.296 & 70.03\% & 16 \\
OCTMNIST & 248,968 & 246,528 & 0.987 & 0.296 & 70.02\% & 16 \\
OrganAMNIST & 250,326 & 247,872 & 0.992 & 0.297 & 70.05\% & 16 \\
\bottomrule
\end{tabular}
\end{table}

The mean source checkpoint is 0.988 MiB and the mean ONNX artifact is 0.296 MiB. The 70.03\% reduction includes graph and container overhead. It should not be interpreted as peak resident-memory reduction.

\subsection{Detailed deployment fidelity}
\label{app:int8_details}
Table~\ref{tab:int8_detail} gives the actual INT8 retention by dataset and training condition. Prediction agreement is computed over all official test images and all ten seeds. The primary-metric delta is the mean absolute checkpoint-level change in Macro-F1 or AUC@0.5. XAI values use the matched intrinsic-map deployment set.

\begin{sidewaystable}[p]
\centering
\caption{Detailed actual ONNX INT8 retention. Lower is better for changed predictions and mean $|\Delta|$; higher is better for agreement, cosine, rank correlation, and top-patch overlap.}
\label{tab:int8_detail}
\scriptsize
\begin{tabular}{llrrrrrr}
\toprule
Dataset & Training condition & Changed / total & Agreement & Mean $|\Delta|$ & Map cosine & Rank $\rho$ & Top-10\% \\
\midrule
BloodMNIST & Adam & 4/34,210 & 99.9883\% & 0.000074 & 0.999960 & 0.994751 & 0.9688 \\
BloodMNIST & Adam + $\mathcal{L}_{\mathrm{attr}}$ & 4/34,210 & 99.9883\% & 0.000082 & 0.999957 & 0.994290 & 0.9675 \\
BloodMNIST & RASO & 3/34,210 & 99.9912\% & 0.000101 & 0.999965 & 0.994616 & 0.9591 \\
PathMNIST & Adam & 14/71,800 & 99.9805\% & 0.000051 & 0.999949 & 0.993345 & 0.9744 \\
PathMNIST & Adam + $\mathcal{L}_{\mathrm{attr}}$ & 12/71,800 & 99.9833\% & 0.000114 & 0.999957 & 0.993766 & 0.9692 \\
PathMNIST & RASO & 9/71,800 & 99.9875\% & 0.000052 & 0.999973 & 0.995187 & 0.9708 \\
BreastMNIST & Adam & 1/1,560 & 99.9359\% & 0.000439 & 0.999970 & 0.998558 & 0.9844 \\
BreastMNIST & Adam + $\mathcal{L}_{\mathrm{attr}}$ & 1/1,560 & 99.9359\% & 0.000439 & 0.999975 & 0.998356 & 0.9844 \\
BreastMNIST & RASO & 0/1,560 & 100.0000\% & 0.000000 & 0.999979 & 0.999309 & 0.9866 \\
PneumoniaMNIST & Adam & 0/6,240 & 100.0000\% & 0.000000 & 0.999949 & 0.988354 & 0.9597 \\
PneumoniaMNIST & Adam + $\mathcal{L}_{\mathrm{attr}}$ & 2/6,240 & 99.9679\% & 0.000085 & 0.999958 & 0.988893 & 0.9591 \\
PneumoniaMNIST & RASO & 2/6,240 & 99.9679\% & 0.000256 & 0.999974 & 0.988819 & 0.9600 \\
DermaMNIST & Adam & 6/20,050 & 99.9701\% & 0.000161 & 0.999947 & 0.987602 & 0.9566 \\
DermaMNIST & Adam + $\mathcal{L}_{\mathrm{attr}}$ & 5/20,050 & 99.9751\% & 0.000019 & 0.999962 & 0.990939 & 0.9594 \\
DermaMNIST & RASO & 3/20,050 & 99.9850\% & 0.000088 & 0.999967 & 0.987713 & 0.9494 \\
OCTMNIST & Adam & 2/10,000 & 99.9800\% & 0.000008 & 0.999960 & 0.997143 & 0.9756 \\
OCTMNIST & Adam + $\mathcal{L}_{\mathrm{attr}}$ & 8/10,000 & 99.9200\% & 0.000276 & 0.999961 & 0.997474 & 0.9797 \\
OCTMNIST & RASO & 2/10,000 & 99.9800\% & 0.000180 & 0.999975 & 0.997072 & 0.9781 \\
OrganAMNIST & Adam & 68/177,780 & 99.9618\% & 0.000135 & 0.999912 & 0.996335 & 0.9650 \\
OrganAMNIST & Adam + $\mathcal{L}_{\mathrm{attr}}$ & 57/177,780 & 99.9679\% & 0.000142 & 0.999889 & 0.997800 & 0.9711 \\
OrganAMNIST & RASO & 37/177,780 & 99.9792\% & 0.000093 & 0.999947 & 0.998024 & 0.9727 \\
\bottomrule
\end{tabular}
\end{sidewaystable}

\subsection{Numerical and runtime audits}
Across all 964,920 source-to-INT8 comparisons, the median per-sample maximum logit error is 0.000203, the 95th percentile is 0.004532, the 99th percentile is 0.014512, the 99.9th percentile is 0.033276, and the maximum is 0.158520. The maximum is an extreme value rather than a typical error; no changed prediction has a source margin exceeding twice its corresponding per-sample maximum logit error.

Every graph has 16 \texttt{MatMulInteger}, 32 \texttt{QuantizeLinear}, 32 \texttt{DequantizeLinear}, and four FP32 attention \texttt{MatMul} nodes. All 210 runtime profiles record integer matrix-multiplication execution. The deployment remains mixed precision because the operations enumerated in Section~3 remain FP32.

\section{Additional Predictive Results and Statistical Tests}
\subsection{Binary secondary metrics}
\label{app:binary}
Table~\ref{tab:binary_secondary} reports conventional probability AUROC in addition to the inherited thresholded metric. The \zach{} rows are FP32; the three \qzach{} rows are actual ONNX INT8 results.

\begin{table}[H]
\centering
\caption{Secondary binary test metrics, mean $\pm$ standard deviation over ten seeds.}
\label{tab:binary_secondary}
\scriptsize
\begin{tabular}{llccc}
\toprule
Dataset & Condition & Probability AUROC & AUC@0.5 & Accuracy \\
\midrule
BreastMNIST & ZACH-ViT + Adam & 0.637 $\pm$ 0.048 & 0.584 $\pm$ 0.045 & 0.535 $\pm$ 0.128 \\
BreastMNIST & qZACH-ViT + Adam & 0.673 $\pm$ 0.043 & 0.619 $\pm$ 0.051 & 0.549 $\pm$ 0.082 \\
BreastMNIST & qZACH-ViT + Adam + $\mathcal{L}_{\mathrm{attr}}$ & 0.671 $\pm$ 0.044 & 0.613 $\pm$ 0.053 & 0.597 $\pm$ 0.104 \\
BreastMNIST & qZACH-ViT + RASO & 0.674 $\pm$ 0.038 & 0.620 $\pm$ 0.045 & 0.572 $\pm$ 0.054 \\
PneumoniaMNIST & ZACH-ViT + Adam & 0.837 $\pm$ 0.017 & 0.738 $\pm$ 0.013 & 0.737 $\pm$ 0.021 \\
PneumoniaMNIST & qZACH-ViT + Adam & 0.836 $\pm$ 0.013 & 0.746 $\pm$ 0.016 & 0.749 $\pm$ 0.027 \\
PneumoniaMNIST & qZACH-ViT + Adam + $\mathcal{L}_{\mathrm{attr}}$ & 0.841 $\pm$ 0.017 & 0.748 $\pm$ 0.018 & 0.743 $\pm$ 0.024 \\
PneumoniaMNIST & qZACH-ViT + RASO & 0.835 $\pm$ 0.020 & 0.745 $\pm$ 0.021 & 0.743 $\pm$ 0.033 \\
\bottomrule
\end{tabular}
\end{table}

\subsection{Paired predictive tests}
\label{app:stats}
Table~\ref{tab:predictive_tests} gives the planned contrasts between the actual ONNX INT8 \qzach{} outputs and the FP32 \zach{} baseline. Holm correction is performed separately for the seven \qzach{} + Adam tests and the seven \qzach{} + \raso{} tests.

\begin{table}[H]
\centering
\caption{Paired Wilcoxon tests across ten seeds. Positive pairs count seed-wise improvements of the deployed INT8 model over the FP32 baseline.}
\label{tab:predictive_tests}
\scriptsize
\begin{tabular}{llrrrr}
\toprule
Dataset & Contrast & Mean delta & Positive pairs & Raw $p$ & Holm $p$ \\
\midrule
BloodMNIST & qZACH + Adam & +0.0478 & 10/10 & 0.0020 & 0.0137 \\
BloodMNIST & qZACH + RASO & +0.0577 & 10/10 & 0.0020 & 0.0137 \\
PathMNIST & qZACH + Adam & +0.0678 & 10/10 & 0.0020 & 0.0137 \\
PathMNIST & qZACH + RASO & +0.0602 & 9/10 & 0.0039 & 0.0195 \\
BreastMNIST & qZACH + Adam & +0.0357 & 6/10 & 0.1934 & 0.3926 \\
BreastMNIST & qZACH + RASO & +0.0360 & 8/10 & 0.1309 & 0.3926 \\
PneumoniaMNIST & qZACH + Adam & +0.0075 & 7/10 & 0.0488 & 0.2441 \\
PneumoniaMNIST & qZACH + RASO & +0.0067 & 7/10 & 0.3750 & 0.7500 \\
DermaMNIST & qZACH + Adam & +0.0138 & 6/10 & 0.3223 & 0.3926 \\
DermaMNIST & qZACH + RASO & +0.0042 & 4/10 & 0.8457 & 0.8457 \\
OCTMNIST & qZACH + Adam & +0.0234 & 7/10 & 0.0840 & 0.3359 \\
OCTMNIST & qZACH + RASO & +0.0356 & 10/10 & 0.0020 & 0.0137 \\
OrganAMNIST & qZACH + Adam & +0.0233 & 7/10 & 0.1309 & 0.3926 \\
OrganAMNIST & qZACH + RASO & +0.0571 & 9/10 & 0.0039 & 0.0195 \\
\bottomrule
\end{tabular}
\end{table}

The global descriptive seed-level summaries are: actual INT8 \qzach{} + Adam, $+0.03132$ with 53 of 70 positive pairs; actual INT8 \qzach{} + Adam + $\Lattr$, $+0.03089$ with 52 positive pairs; and actual INT8 \qzach{} + \raso{}, $+0.03677$ with 57 positive pairs. These 70 values combine different datasets and are not used as the sole inferential basis.

\section{XAI Compatibility, Metrics, and Sanity Checks}
\subsection{Compatibility audit}
Every implemented method returned a finite normalized 196-patch map on a trained \qzach{} + \raso{} checkpoint. Token Grad-CAM treats the final token grid as the feature map. Attention Rollout-ZT and Gradient Attention Rollout-ZT average final-query relevance because the architecture uses global average pooling and no classification token. Exact transformer LRP is excluded rather than approximated because no relevance-conservation derivation was implemented for the custom quantization-aware training modules and recursive evidence fusion.

\subsection{Intrinsic-XAI paired tests}
\label{app:xai_stats}
Table~\ref{tab:xai_stats} compares \raso{} with Adam plus the same attribution loss after averaging examples within each of 70 dataset-seed checkpoints. Holm correction covers ten metrics.

\begin{table}[H]
\centering
\caption{Paired intrinsic-XAI comparison: \qzach{} + \raso{} minus \qzach{} + Adam + $\Lattr$.}
\label{tab:xai_stats}
\scriptsize
\begin{tabular}{llrrr}
\toprule
Metric & Better direction & Mean delta & Raw $p$ & Holm $p$ \\
\midrule
Deletion AUC & lower & -0.00432 & 0.12449 & 0.65528 \\
Insertion AUC & higher & -0.00324 & 0.35974 & 1.00000 \\
Comprehensiveness & higher & -0.01181 & 0.09361 & 0.65528 \\
Sufficiency error & lower & -0.02998 & 0.00138 & 0.01106 \\
SaCo & higher & -0.00790 & 0.81719 & 1.00000 \\
Occlusion rank correlation & higher & +0.02172 & 0.92772 & 1.00000 \\
Noise cosine & higher & +0.00177 & 0.00000 & 0.00002 \\
Noise rank correlation & higher & +0.00868 & 0.09476 & 0.65528 \\
Noise top-10\% overlap & higher & -0.00306 & 0.51407 & 1.00000 \\
Noise JS & lower & -0.00055 & 0.00000 & 0.00000 \\
\bottomrule
\end{tabular}
\end{table}

\subsection{Parameter-randomization checks}
\label{app:sanity}
The expected top-10\% overlap for independent rankings is approximately 0.10. Low rank correlation and chance-level overlap therefore provide evidence of parameter dependence even when cosine is inflated by smooth maps.

\begin{table}[H]
\centering
\caption{Similarity between explanations before and after full model-parameter randomization. Lower cosine, rank, and overlap and higher JS indicate stronger sensitivity to learned parameters.}
\label{tab:sanity}
\scriptsize
\begin{tabular}{lrrrrr}
\toprule
Method & $n$ & Cosine & Rank $\rho$ & Top-10\% overlap & JS \\
\midrule
\textbf{qZACH intrinsic} & 28 & 0.963 & -0.163 & 0.109 & 0.0102 \\
Integrated Gradients & 28 & 0.628 & 0.086 & 0.225 & 0.0890 \\
Token Grad-CAM & 28 & 0.919 & 0.028 & 0.189 & 0.0373 \\
Attention Rollout-ZT & 28 & 0.861 & 0.158 & 0.259 & 0.0210 \\
Gradient Attention Rollout-ZT & 28 & 1.000 & -0.130 & 0.141 & 0.0000 \\
RISE & 28 & 0.999 & -0.076 & 0.096 & 0.0002 \\
\bottomrule
\end{tabular}
\end{table}

The high cosine values of several methods should not be interpreted in isolation. For \qzach{} intrinsic attribution, cosine is 0.963, while rank correlation is $-0.163$ and top-10\% overlap is 0.109. Gradient Attention Rollout-ZT and RISE have near-unit cosine and tiny JS but rank and top-patch changes closer to chance. Integrated Gradients is the most clearly parameter-sensitive under all four measures in this audit.

\section{Additional Qualitative Comparisons}
The following panels show one fixed seed-3 example per remaining dataset. They are not selected as best cases and include errors. Each panel compares the same \qzach{} + \raso{} prediction across intrinsic and post-hoc methods.

\begin{sidewaysfigure}[p]
\centering
\includegraphics[width=0.97\textheight]{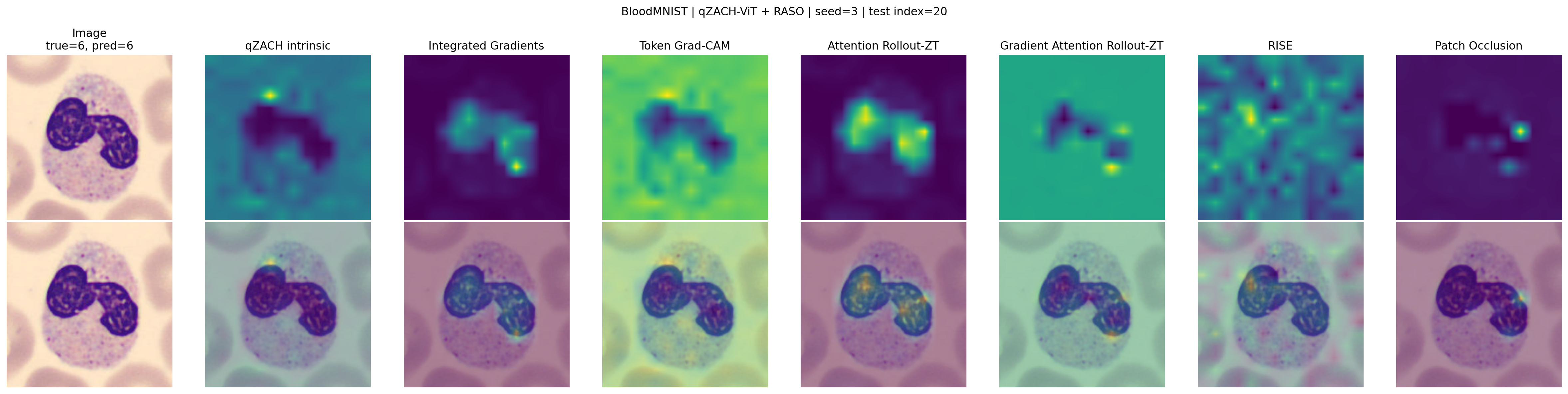}
\caption{BloodMNIST matched XAI methods.}
\end{sidewaysfigure}

\begin{sidewaysfigure}[p]
\centering
\includegraphics[width=0.97\textheight]{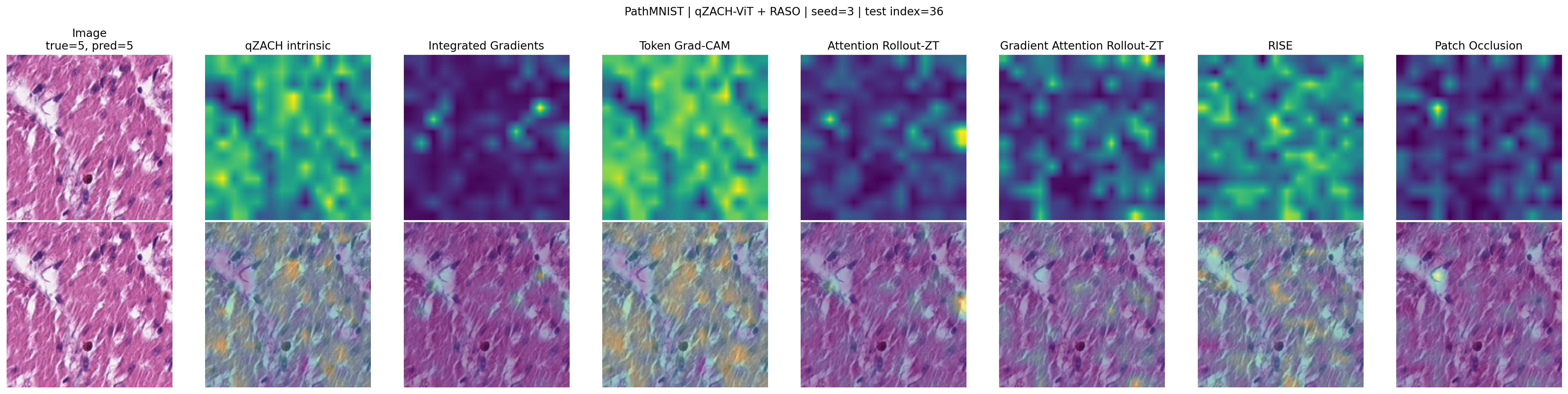}
\caption{PathMNIST matched XAI methods.}
\end{sidewaysfigure}

\begin{sidewaysfigure}[p]
\centering
\includegraphics[width=0.97\textheight]{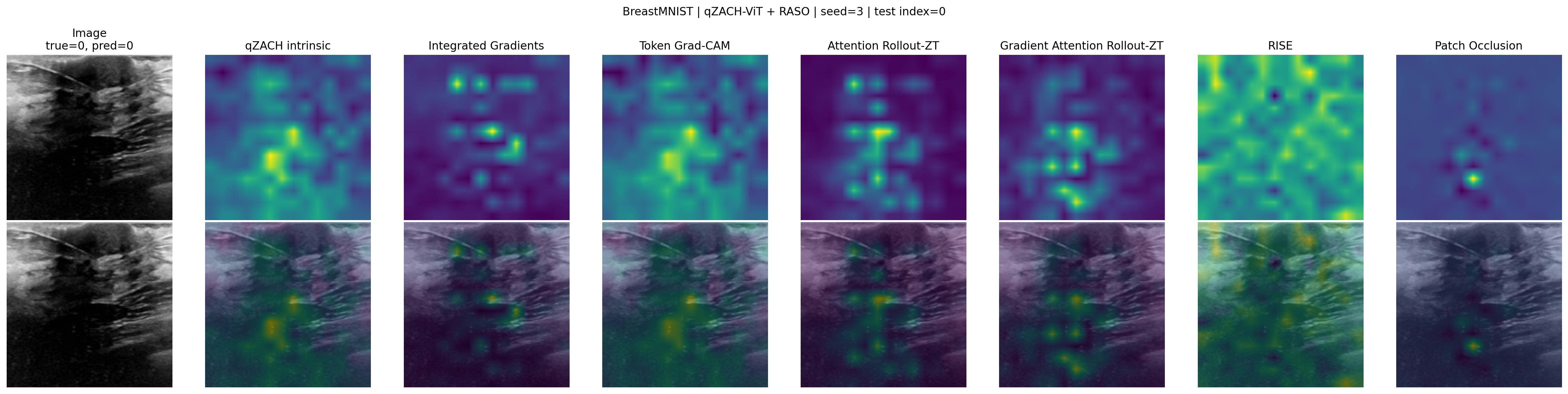}
\caption{BreastMNIST matched XAI methods.}
\end{sidewaysfigure}

\begin{sidewaysfigure}[p]
\centering
\includegraphics[width=0.97\textheight]{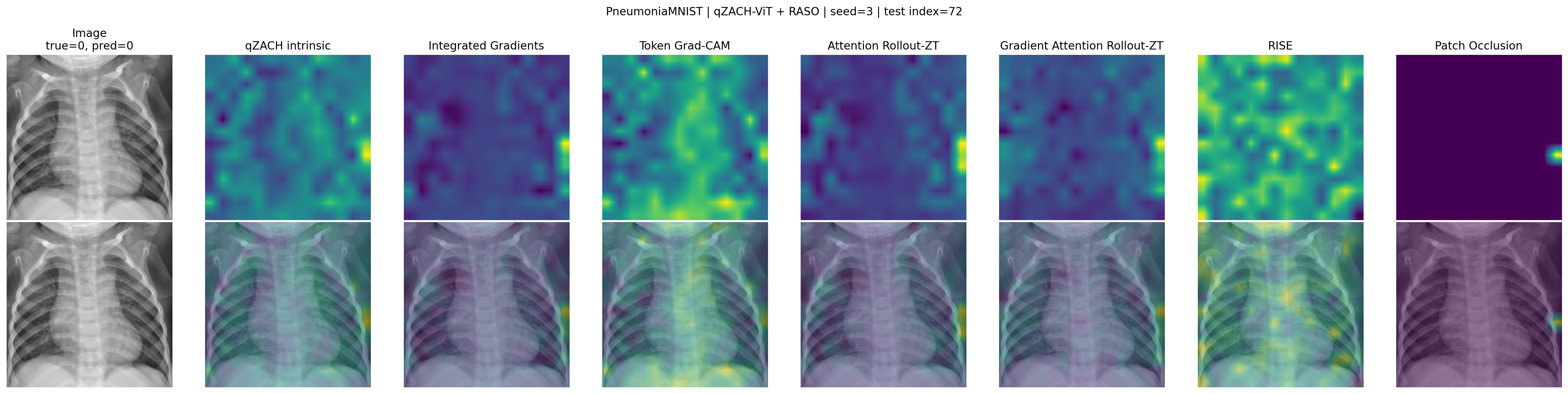}
\caption{PneumoniaMNIST matched XAI methods.}
\end{sidewaysfigure}

\begin{sidewaysfigure}[p]
\centering
\includegraphics[width=0.97\textheight]{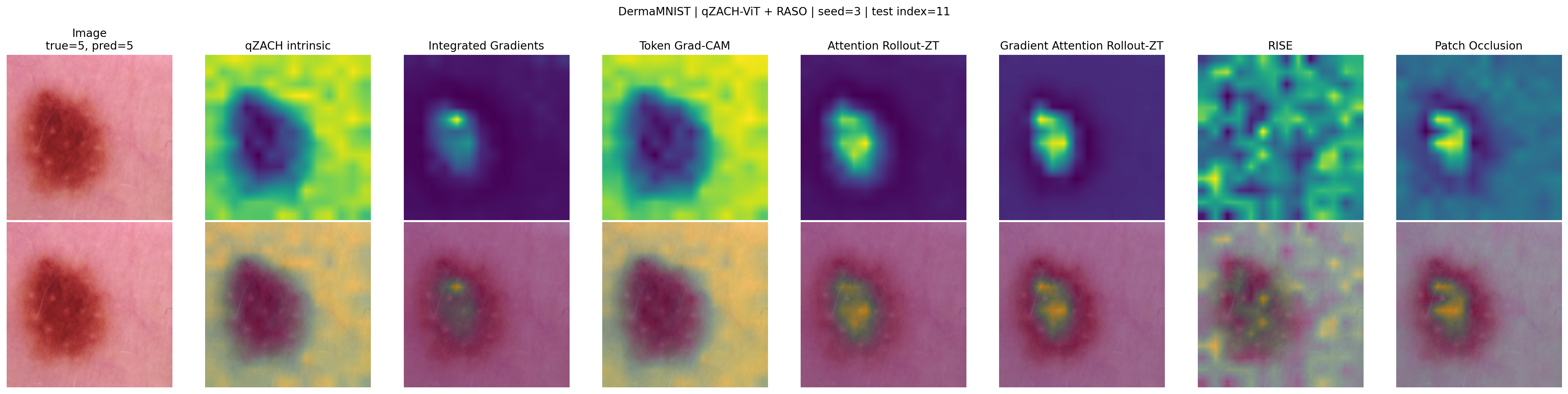}
\caption{DermaMNIST matched XAI methods.}
\end{sidewaysfigure}

\begin{sidewaysfigure}[p]
\centering
\includegraphics[width=0.97\textheight]{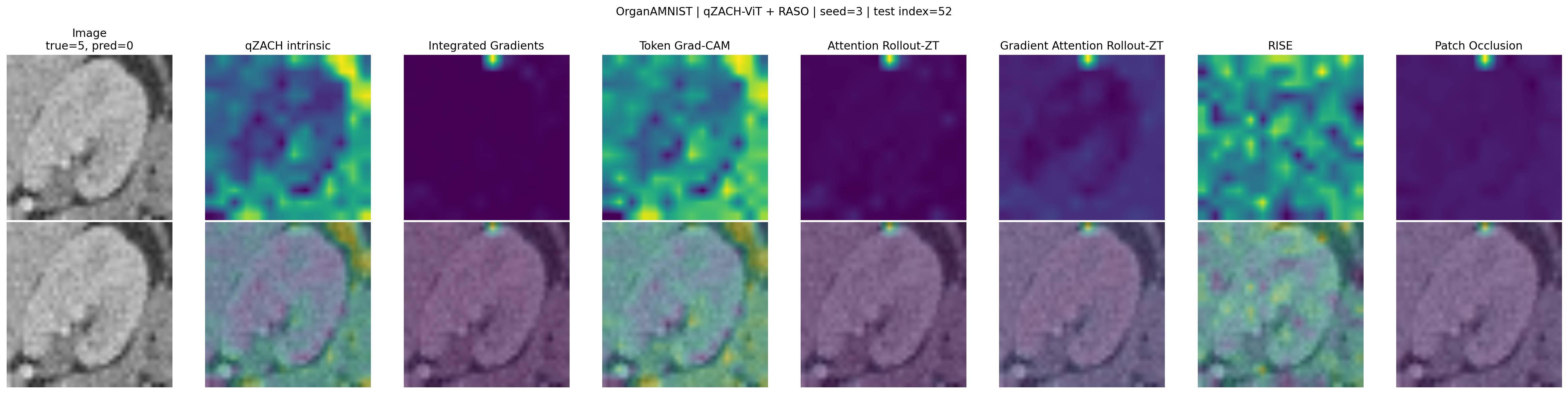}
\caption{OrganAMNIST matched XAI methods.}
\end{sidewaysfigure}

\end{document}